\pdfoutput=1

\documentclass[journal]{IEEEtran}

\usepackage{xcolor,soul,framed} 

\colorlet{shadecolor}{yellow}
\usepackage[pdftex]{graphicx}
\usepackage{subfigure}
\usepackage{caption}
\usepackage{graphicx}
\usepackage{float} 
\usepackage{caption}
\usepackage{booktabs}
\usepackage{multirow}
\graphicspath{{../pdf/}{../jpeg/}}
\DeclareGraphicsExtensions{.pdf,.jpeg,.png}

\usepackage[cmex10]{amsmath}
\usepackage{array}
\usepackage{mdwmath}
\usepackage{mdwtab}
\usepackage{eqparbox}
\usepackage{url}
\usepackage{setspace}

\usepackage{hyperref} 
\hypersetup{
    colorlinks=true,
    linkcolor=blue,
    anchorcolor=blue,
    citecolor=blue
} 
\usepackage{hyperref}
\usepackage{fbox}

\usepackage{booktabs, makecell, multicol}

\begin{document}
\bstctlcite{IEEEexample:BSTcontrol}
    \title{Bridging the Gap Between Theory and Practice: Benchmarking Transfer Evolutionary Optimization
    }
% \author{\IEEEauthorblockN{Anonymous Authors}}
    \author{
        Yaqing~Hou,
        Wenqiang~Ma,
        Abhishek~Gupta,
        Kavitesh~Kumar~Bali,
        Hongwei~Ge, 
        Qiang~Zhang,
        Carlos A. Coello Coello~\IEEEmembership{Fellow,~IEEE}
        and Yew-Soon~Ong,~\IEEEmembership{Fellow,~IEEE}

\thanks{This work was supported in part by the National Key Research and Development Program of China (No. 2021ZD0112400), the National Natural Science Foundation of China under Grant 62372081, the Basque Government through the BERC 2022-2025 program, the Spanish Ministry of Economy and Competitiveness MINECO: BCAM Severo Ochoa excellence accreditation SEV-2017-0718, the Young Elite Scientists Sponsorship Program by CAST under Grant 2022QNRC001, and the NSFC-Liaoning Province United Foundation under Grant U1908214, the 111 Project, No.D23006.}% <-this % stops a space
\thanks{Yaqing Hou, Wenqiang Ma, Hongwei Ge, and Qiang Zhang are with the School of Computer Science and Technology, Dalian University of Technology (DLUT), China 116024 (E-mails: houyq@dlut.edu.cn; ranyi@mail.dlut.edu.cn; gehw@dlut.edu.cn; zhangq@dlut.edu.cn).}

\thanks{Abhishek Gupta is with the School of Mechanical Sciences, Indian Institute of Technology (IIT) Goa, India (E-mail: abhi.nitr2010@gmail.com).}

\thanks{Kavitesh~Kumar~Bali is with the Centre for Frontier AI Research (CFAR), Agency for Science, Technology and Research (A*star), Singapore (E-mail: bali.kavitesh@gmail.com).}

\thanks{Carlos A. Coello Coello is with the Departamento de Computación, CINVESTAV-IPN, Mexico City, Mexico. This work is part of his sabbatical leave at the Basque Center for Applied Mathematics (BCAM) \& Ikerbasque, Spain (e-mail: carlos.coellocoello@cinvestav.mx).}

\thanks{Yew-Soon Ong is the Chief Artificial Intelligence Scientist with the Agency for Science, Technology and Research (A*STAR), Singapore, and is also with the Data Science and Artificial Intelligence Research Centre, School of Computer Science and Engineering, Nanyang Technological University (NTU), Singapore (email: asysong@ntu.edu.sg).}
}

\maketitle

\begin{abstract}
In recent years, the field of Transfer Evolutionary Optimization (TrEO) has witnessed substantial growth, fueled by the realization of its profound impact on solving complex problems. Numerous algorithms have emerged to address the challenges posed by transferring knowledge between tasks. However, the recently highlighted ``no free lunch theorem'' in transfer optimization clarifies that no single algorithm reigns supreme across diverse problem types. This paper addresses this conundrum by adopting a benchmarking approach to evaluate the performance of various TrEO algorithms in realistic scenarios. Despite the growing methodological focus on transfer optimization, existing benchmark problems often fall short due to inadequate design, predominantly featuring synthetic problems that lack real-world relevance. This paper pioneers a practical TrEO benchmark suite, integrating problems from the literature categorized based on the three essential aspects of Big Source Task-Instances: volume, variety, and velocity. Our primary objective is to provide a comprehensive analysis of existing TrEO algorithms and pave the way for the development of new approaches to tackle practical challenges. By introducing realistic benchmarks that embody the three dimensions of volume, variety, and velocity, we aim to foster a deeper understanding of algorithmic performance in the face of diverse and complex transfer scenarios. This benchmark suite is poised to serve as a valuable resource for researchers, facilitating the refinement and advancement of TrEO algorithms in the pursuit of solving real-world problems.
\end{abstract}

\begin{IEEEkeywords}
Optimization experience, transfer evolutionary optimization, no free lunch theorem,  practical test problems, benchmark suite.
\end{IEEEkeywords}

\IEEEpeerreviewmaketitle

\section{Introduction}

Evolutionary algorithms (EAs) have gained widespread applications in solving a diverse range of optimization problems due to their versatility and ease of use~\cite{ea-applications}. These algorithms find utility in industrial applications, including resource scheduling~\cite{resource-constrained},~\cite{Resource-Investment-Project-Scheduling}, mechanical design~\cite{industrial-design},~\cite{Satellite-Module-Layout-Design}, cyber security~\cite{cyber-physical-systems},~\cite{cybersecurity},~\cite{cyber-attack-prevention}, vehicle routing~\cite{Vehicle-Routing-Problem-With-Time-Windows},~\cite{Two-Echelon-Vehicle-Routing}, and machine learning~\cite{Evolutionary-Machine-Learning},~\cite{Evolutionary-Automated-Machine-Learning}. However, a limitation of conventional evolutionary optimization methods is their {\em tabula rasa} design, by which they often start an optimization process from scratch assuming little prior knowledge about solutions to the problem at hand.  It is crucial to recognize that real-world problems seldom exist in isolation~\cite{CIGAR}; they often share commonalities with numerous instances solved in the past (e.g., within a specific industry)~\cite{air5}. This realization has spurred a growing interest in the development of efficient optimizers capable of enhancing their performance by automatically transferring and reusing acquired knowledge across related problems~\cite{insights}.

In particular, transfer evolutionary optimization (TrEO) has emerged as a computational intelligence paradigm that facilitates knowledge transfer in evolutionary computation from previously solved \emph{source} tasks to address a new \emph{target} task of interest~\cite{insights}. In recent years, numerous TrEO algorithms have showcased their efficacy across various domains, emphasizing the potential for specialization in evolutionary optimization~\cite{revies-on-mto,half-a-dozen}. Despite these advancements, the recently proven ``no free lunch theorem'' (NFLT) in transfer optimization~\cite{EKT} highlights a critical observation - no single algorithm universally outperforms others across diverse problem types. A comprehensive literature review, exemplified by studies like~\cite{EKT}, substantiates the NFLT's validity, revealing no algorithm's supremacy over others across numerous synthetic problems. This phenomenon is conjectured to become even more apparent in practical scenarios, where a multitude of generated source tasks often exhibit higher diversity and complexity. 

Existing TrEO studies often focus on empirical analyses of optimization performance using synthetic benchmark functions in idealized settings~\cite{multitasking-single-benchmark, multitasking-multi-benchmark, cec2020benchmark}. Many of these functions are perceived as disconnected from reality. A recent paper has offered proof of faster convergence through transfer in the surrogate-assisted optimization setting~\cite{ExTrEMO}. However, aside from a handful of such work, the general shortage of formal results offering performance guarantees for TrEO, coupled with the need for better evaluation and benchmarking~\cite{EKT}, poses challenges in understanding how the proposed algorithms would perform in real applications. Without appropriate benchmark problems, a comprehensive assessment of TrEO's efficacy in coping with the implications of the NFLT in real-world situations is difficult.

To address this limitation, this paper adopts a benchmarking approach to evaluate the performance of several TrEO algorithms in realistic scenarios. Our objective is to gain a deeper understanding of algorithmic performance when confronted with diverse and complex transfer scenarios, each possessing its unique strengths and weaknesses. The paper pioneers a practical TrEO benchmark suite, integrating problems from the literature categorized based on three postulated \emph{V's} of \emph{Big Source Task-Instances}: \emph{Big Volume} (given a static but substantial number of source tasks), \emph{Big Variety} (under a wide range of source tasks with varying elements of optimization problems), and \emph{Big Velocity} (for efficient transfer optimization in time-sensitive environments). By introducing realistic benchmarks embodying the three dimensions of volume, variety, and velocity, we aim to provide a comprehensive analysis of existing TrEO algorithms and foster a deeper understanding of algorithmic performance in the face of diverse and complex transfer scenarios.

In practice, the increasing \emph{volume} of source tasks presents obstacles such as memory cost and the computational burden of selecting relevant information with prospects for positive transfers~\cite{coping-with-big-data-MAB-AMTEA}. The \emph{variety} of possible characteristics of source task-instances, in terms of input features, dimensionality, objective function landscapes, etc., points to the importance of enhancing the generality of TrEO methods in exploiting transferred knowledge in optimization problems~\cite{insights}. What's more, in online or lifelong learning scenarios with growing \emph{velocity} of source information streams, there is a pressing need for better learning agility to quickly and effectively utilize time-sensitive ``fresh'' information, given the possible sparsity of relevant sources to the target task~\cite{sTrEO}. 

The organization of the rest of this paper is as follows. 
Section~\ref{background} presents a brief background of TrEO. Section~\ref{benchmark} analyzes the characteristics of practical problems related to Big Source Task-Instances and categorizes the problems included in our test suite. Section~\ref{probdescription} gives a description and a measure of similarity between tasks of the three proposed benchmark problems. Section~\ref{baseline} offers comprehensive experimental details and presents baseline results derived from several mainstream algorithms, providing empirical support for the ``no free lunch theorem''. Section~\ref{conclusion} concludes the paper.

\section{Background}
\label{background}

This section lays the groundwork for transfer optimization. It introduces the basic problem formulation and discusses several state-of-the-art transfer evolutionary algorithms.

\subsection {Preliminaries on Transfer Optimization}

In a canonical case with no transfer, an optimization problem/task $\mathcal{T}$ with an objective function $f(x)$ to be maximized can be stated as: 
\begin{equation}\label{transfer_optimization_formulation}
    \max\limits_{x \in \mathcal{X}} f(x),
\end{equation}
where $\mathcal{X}$ represents the search space of the problem and $x \in \mathcal{X}$ denotes a candidate solution~\cite{domain-adaptation}\cite{memetic-computation:}. $f(x)$ may be a black box whose mathematical form and derivatives are unknown/inaccessible.
A run of an optimization algorithm could then be called successful if the set of solutions $X^*$ evaluated satisfy the following condition:
\begin{equation}\label{successful_transfer_optimization}
    f(x) \ge f^* - \epsilon,\ x \in X^* .
\end{equation}
Here, $f^*$ represents the true global optimum of the objective function, and $\epsilon$ is a small (positive) tolerance threshold.

Standard optimization methods often suffer from high computational costs as they begin every optimization task from scratch without dedicated mechanisms for utilizing experiential knowledge. In contrast, \emph{transfer optimization} harnesses the power of knowledge gained from previously solved tasks, leading to faster convergence and enhanced performance.

In the sequential transfer optimization setting, we consider a collection of $K-1$ previously solved source problems denoted as $\mathcal{T}_1, \mathcal{T}_2, ..., \mathcal{T}_{K-1}$. 
Each task is associated with an objective function, represented as $f_1, f_2, ..., f_{K-1}$. In the context of probabilistic evolutionary search, let a population's underlying probability distribution model be denoted as $p(x)$. Then, for those previously optimized source tasks, their respective optimized search distributions, {$p_1^*(x)$, $p_2^*(x)$, ..., $p_{K-1}^*(x)$}, are seen as building blocks of knowledge available for reuse. For the specific target task $\mathcal{T_K}$ with an objective function $f_K$, let $p_K(x)$ be the probabilistic model of the population in $\mathcal{T_K}$. 

Under this scenario, we can reformulate equation~(\ref{transfer_optimization_formulation}) in an information-geometric form~\cite{information-geometric} where it is possible to optimize the population distribution of the target task $\mathcal{T}_K$ while leveraging the source priors as~\cite{domain-adaptation},
\begin{equation}\label{sequential_transfer_formulation}
    % \begin{tiny}
        \begin{aligned}
        \mathop{max}_{w_1, w_2, ..., w_K, p_K(x)}& \int_{\mathcal{X}}f_K(x)\\ 
        \cdot &\left[ \sum_{s=1}^{K-1} w_s \cdot p_{M_s}^*(x) + w_K \cdot p_K(x)\right] \cdot \mathrm{d}x, \\
        s.t.,\ \sum_{i=1}^K w_i = &
        1\ and\ w_i \ge 0,\forall i \in \left\{1, ..., K-1, K\right\} .\\  
        \end{aligned}
        % \end{tiny}
\end{equation}

Here, $x$ represents a candidate solution for the target task $\mathcal{T}_K$. $M_s$ represents a mapping function designed to transform solutions from the $s^{th}$ source to the target, enabling comprehensive utilization of stored knowledge derived from the source tasks. $p_{M_s}^*(x)$ signifies the adaptation of $p_s^*(x)$ through the mapping function $M_s$. $w_1, \ldots, w_{K-1}, w_K$ are transfer coefficients that describe the probabilistic mixture model $\left[ \sum_{s=1}^{K-1} w_s \cdot p_{M_s}^*(x) + w_K \cdot p_K(x)\right]$. These transfer coefficients play a crucial role in determining the extent of knowledge (i.e., the number of solutions) transferred from the various components of the mixture model to the target. If a particular source is highly correlated with the target task, then achieving a high transfer coefficient corresponding to that source will facilitate rapid target optimization. 

However, fine-tuning the transfer coefficients is nontrivial due to the absence of precise knowledge of inter-task correlations, especially in black-box optimization settings. The repercussions of incorrect transfer coefficient assignments are two-fold. On one hand, allocating an unreasonably high transfer coefficient to an unrelated source model introduces useless or even harmful solutions into the target's search space, thus potentially impeding overall search performance. Conversely, the inability to assign an appropriately high value to a related source model hampers the transfer of valuable solutions, limiting potential optimization enhancements for the target task. Therefore, it becomes crucial for a transfer optimization algorithm to accurately calibrate the mixture transfer coefficients of equation~(\ref{sequential_transfer_formulation}), leveraging solution-evaluation data generated during the target optimization search.

\begin{figure}
    \centering
    \includegraphics[width=0.5\textwidth]{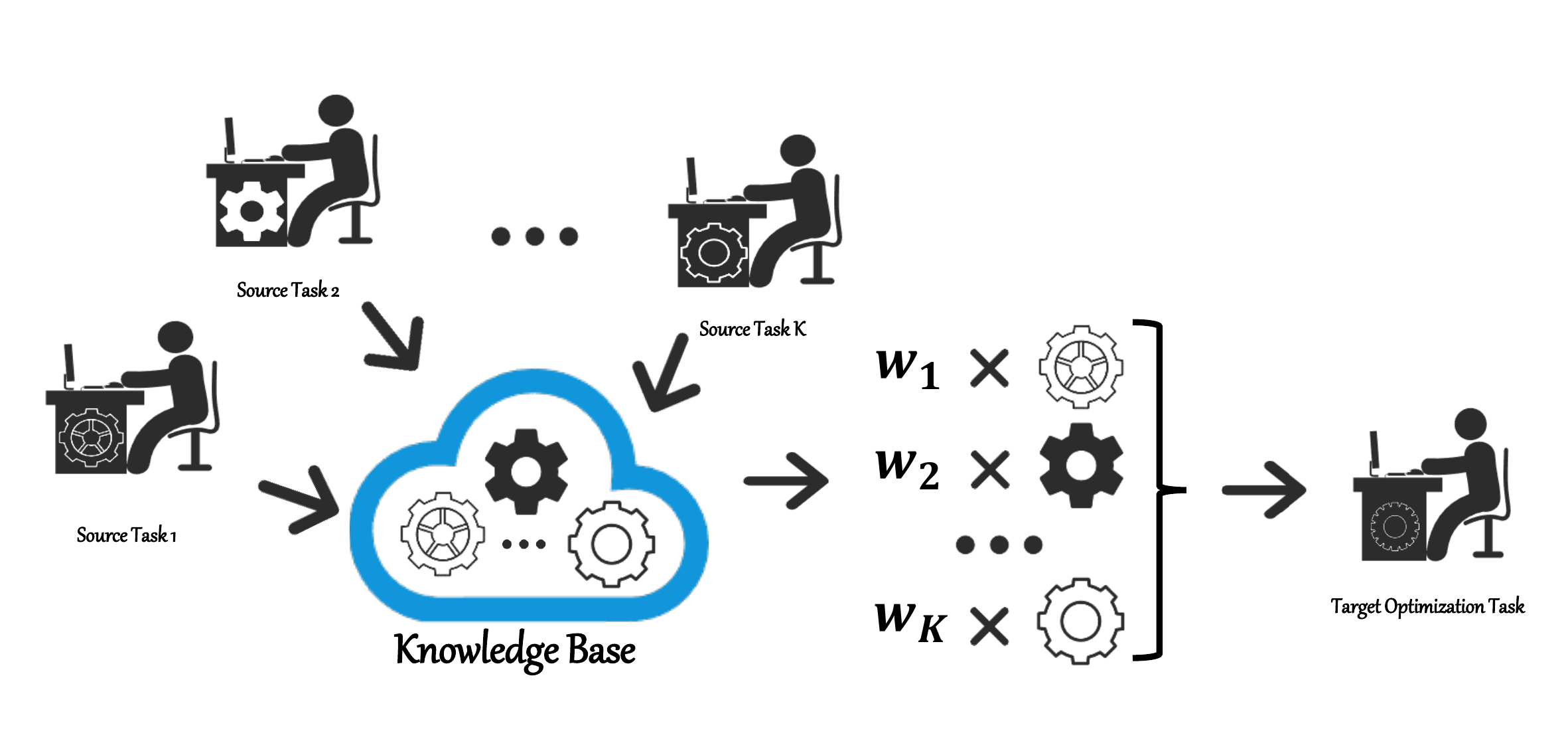}
    \caption{%Illustration of sequential transfer optimization: In sequential transfer, all the knowledge from pre-optimized tasks (called source tasks, in this case, running, high jumping and long jumping) were extracted and stored in the knowledge base for later solving target task (in this case, hurdling).
    In sequential transfer optimization, knowledge from pre-optimized tasks is extracted and stored in a knowledge base. This accumulated knowledge is later utilized to solve the target task.}
    \label{sequential-transfer}
    
\end{figure}

\subsection {State-of-the-Art Transfer EAs}
\label{background-sota}

Transfer evolutionary optimization (TrEO) is a promising approach for reusing source priors for efficient target problem-solving~\cite{insights, ETO}. 
Recently, a variety of \textit{TrEO} algorithms 
have been proposed in a multitude of areas, including neuro-evolution~\cite{neuro-evolution}, multi-objective optimization~\cite{multiproblem}, objective-heterogeneous optimization~\cite{objective-heterogeneous}, dynamic optimization~\cite{population-based, direct-memory-schemes}, combinatorial optimization 
\cite{memetic-search, memes-as-building-blocks}, symbolic regression~\cite{transfer-learning-in-gp,further-investigation}, learning classifier systems~\cite{extracting},~\cite{reusing-building-blocks}, evolutionary robotics~\cite{evolutionary-robotics}, image classification~\cite{cross-domain-reuse} and reinforcement learning~\cite{frame-correlation}. Many of these algorithms align with the concept described in equation~(\ref{sequential_transfer_formulation}) and generally exhibit superior performance compared to those approaches that do not utilize transfer optimization~\cite{genetic-transfer-learning}. 

In terms of implementation of the concept, there are two main strategies to extract knowledge from prior source tasks\footnote{Table~\ref{algorithms-categories} provides additional categorization details for existing TrEO algorithms.}: model-free transfer and model-based transfer. While this paper does not offer a comprehensive methodological review of each strategy, it does spotlight several noteworthy state-of-the-art examples for illustration purposes.

\begin{table}[htbp]
    \newcommand{\tabincell}[2]{\begin{tabular}{@{}#1@{}}#2\end{tabular}}
    \centering
    \caption{Representative transfer evolutionary algorithms.}
    \setlength{\tabcolsep}{0.8mm}{
        \small
        \begin{tabular}{cccc}
            \toprule
            \makecell[c]{Algorithms} & \makecell[c]{Probabilistic\\Model-based?} & \makecell[c]{Transformation-/\\Mapping-based?} & \makecell[c]{Adaptive \\Transfer?} \\
            \midrule
            CAMA-M~\cite{memes-as-building-blocks} & no & yes & yes \\  % (2015)
            \hline
            MIPBIL~\cite{direct-memory-schemes} & no & no & no \\  % (2016)
            \hline
            KT~\cite{further-investigation}  & no & no & yes \\  % (2016)
            \hline
            TEMO-MPS~\cite{multiproblem} & yes & no & yes \\  % (2017)
            \hline
            TLGP-criptor~\cite{cross-domain-reuse} & no & no & yes \\  % (2017)
            \hline
            AMTEA~\cite{curbing-AMTEA} & yes & no & yes \\  % (2018)
            \hline
            MFEA-DV~\cite{MFEA-DV} & no & no & no \\  % (2019)
            \hline
            MSSTO~\cite{MSSTO} & no & yes & yes \\  % (2020)
            \hline
            tNES~\cite{neuro-evolution} & yes & no & yes\\  % (2021)
            \hline
            DVA-ESTO~\cite{objective-heterogeneous} & no & yes & no \\  % (2021)
            \hline
            KAES~\cite{KAES} & no & yes & no \\  % (2021)
            \hline
            AT-MFEA~\cite{AT-MFEA}& no & yes& no \\  % (2022)
            \hline
            sTrEO~\cite{sTrEO} & yes & no & yes \\  % (2022)
            \hline
            MSTL-DMOEA~\cite{MSTL-DMOEA} & no & yes & yes \\  % (2022)
            \hline
            EKT~\cite{EKT} & no & no & no \\  % (2023)
            \bottomrule
        \end{tabular}
        }
    \label{algorithms-categories}
\end{table}

(1) \textit{Model-Free Transfer}: The objective of this strategy is to transfer supplementary information linked to solutions into the target task. Solutions from analogous source tasks previously solved are archived in a knowledge base. When addressing the target task, pertinent auxiliary details are extracted from these stored solutions and introduced into the optimization procedure. Several model-free transfer optimization algorithms have been proposed in the literature, including CIGAR~\cite{CIGAR, CIGAR-application}, MFEA-DV~\cite{MFEA-DV}, MSSTO~\cite{MSSTO}, KAES~\cite{KAES}, AT-MFEA~\cite{AT-MFEA}, and MSTL-DMOEA~\cite{MSTL-DMOEA}, among others. For instance, CIGAR combines Genetic Algorithms (GA) and case-based memory to continuously improve performance on sets of similar problems. It periodically injects suitable intermediate solutions from similar previously solved source tasks into the GA's population, rather than starting the optimization process anew for the target task. In other works such as~\cite{minimalistic-attack} and~\cite{frame-correlation}, the approach is simpler, where the best solution from a similar previously solved source task is incorporated to initialize the GA's initial population, and the rest of the procedure follows a standard GA.
     
(2) \textit{Model-Based Transfer}: This strategy leverages probabilistic models or other models generated from candidate solutions of previously solved source tasks to reuse prior knowledge. The models built from candidate solutions of solved source tasks are stored and utilized for subsequent optimization of similar target tasks. In the field of model-based sequential transfer methods, several representative algorithms include KBOA~\cite{KBOA}, MOI-MBO~\cite{MOI-MBO}, AMTEA~\cite{curbing-AMTEA}, MAB-AMTEA~\cite{coping-with-big-data-MAB-AMTEA}, DVA-ESTO~\cite{objective-heterogeneous}, and sTrEO~\cite{sTrEO}. For example, AMTEA is an adaptive model-based transfer evolutionary algorithm designed to minimize the risk of negative transfer. However, its source-target similarity capture exhibits limitations as the number of source tasks increases rapidly. To address this issue, the Multi-Armed Bandit (MAB) theory was recently introduced to the AMTEA. However, MAB-AMTEA selects only one source to extract knowledge, which may be challenging when useful sources represent only a small percentage of the overall source tasks. In contrast, sTrEO focuses on both scalability against a growing number of source instances and online learning agility against sparsity of highly related source tasks to the target task. Experimental results show that sTrEO achieves superior performance compared to the aforementioned algorithms.
     
\section{Benchmark Categorization}
\label{benchmark}

In this section, we analyze the characteristics of transfer optimization of big task instances and present the benchmarking problem-suite categories. 

The rise of Big Source Task-Instances and the increasing number of potentially similar historical source tasks pose challenges in efficiently reusing knowledge for TrEO. We investigate three fundamental characteristics of TrEO problems related to \textit{Big Volume}, \textit{Big Variety}, and \textit{Big Velocity}, which shape the performance and effectiveness of TrEO algorithms in real-world problem-solving scenarios. These characteristics play a crucial role in determining the scalability, adaptability, and efficiency of TrEO algorithms in handling large and diverse source tasks, as well as in addressing transient and rapidly changing optimization landscapes. By categorizing our benchmarking test suite based on these characteristics, we aim to provide a comprehensive evaluation of TrEO algorithms' capabilities and limitations across different problem domains and complexities.

\subsection{Big Volume of Source Task-Instances}

\begin{center}
\fbox{
  \parbox{0.45\textwidth}{
    \textbf{Definition 1:} A Big Volume of Source Task-Instances: a static but substantial number of previously solved source tasks.
  }
}
\end{center}

The masses of source tasks could contain potentially useful knowledge for optimizing a target task. TrEO with a big volume of source task instances poses several challenges in practical optimization scenarios. 
One major challenge is the increased computational cost required to handle large volumes of source data~\cite{coping-with-big-data-MAB-AMTEA}. 
Furthermore, the transfer of knowledge from a multitude of related problems becomes more difficult, and the risk of negative transfer increases. Identifying relevant and transferable knowledge among the large pool of source tasks becomes a critical task~\cite{sTrEO}.
The issue is compounded when a significant proportion of the source tasks are not relevant to the target optimization task. This necessitates the development of specialized techniques and the use of powerful computational resources to effectively tackle the challenges posed by big volumes of source tasks. For instance, in the large-scale cloud-based platforms of today, online services catering to thousands
of diverse clients worldwide have emerged. Locating the relevant source priors among these large volumes of ``pre-solved'' cases could have a significant impact on enhancing service efficiency as well as the quality of solutions recommended to any new target client.

\subsection{Big Variety of Source Task-Instances}
\begin{center}
\fbox{
  \parbox{0.45\textwidth}{
    \textbf{Definition 2:} A Big Variety of Source Task-Instances: a large range of source tasks with different elements of optimization problems, including differing dimensionalities and solution representation of variables, heterogeneity of the search space, differing objectives and constraints, etc.
  }
}
\end{center}

In a significant body of TrEO studies, researchers often assumed the configuration of identical optimization elements across source and target tasks~\cite{domain-adaptation}. Therefore, the well-optimized knowledge extracted from source tasks can positively direct the search process of target tasks. In real-world scenarios, the mismatch in optimization elements between source and target tasks often occurs. However, this does not necessarily imply that the unrelatedness across tasks and their true relationship may be concealed.
This necessitates finding appropriate transformations or mappings of search spaces between mismatched optimization elements to increase the overlap in solution distributions and uncover latent similarities or connections between tasks~\cite{solution-representation}.
To illustrate this, consider the example of a planar robotic arm task, where the goal is to reach a pre-specified point with a robotic arm possessing specific characteristics. In this scenario, the source tasks may exhibit variations in optimization elements, such as dimensionalities and robot arm length, leading to differences in variable and objective features. Consequently, knowledge derived from source tasks needs to undergo a mapping or a transformation before it can be effectively utilized, as the variations in the optimization elements create different domains for each task. Therefore, addressing the wide variety of source task-instances demands the development of transfer optimization methods that can adapt to transfer between tasks with diverse optimization elements by mapping or transforming knowledge from different domains for efficient knowledge transfer. 

\subsection{Big Velocity of Source Task-Instances}
\begin{center}
\fbox{
  \parbox{0.45\textwidth}{
    \textbf{Definition 3:} A Big Velocity of Source Task-Instances: the high speed of optimizing target problems towards a certain level of fitness performance (often by leveraging knowledge from a considerable number of source tasks) given limited time and computational power.
  }
}
\end{center}

It is a requirement for efficient transfer optimization in time-sensitive environments and poses a significant challenge in practical transfer optimization problems. As the size of the source data increases, the algorithm's analysis time also grows. In many real-world optimization scenarios, the need to find an optimal solution within the shortest possible time is crucial~\cite{to-handle-big-data}. Furthermore, some practical problems demand algorithms to achieve satisfactory objective values within strict time constraints as well~\cite{real-time-optimization}.
For instance, consider the context of attacking a pre-trained policy in a sequential decision-making setting, where the optimization algorithm must add perturbations to an agent's observations to deceive it into altering its actions. Such attacks can be seen as a form of defense against an offensive adversary. Here, time is limited, and the algorithm must launch a successful attack within a tight time window before the key frame vanishes. 
In such contexts, time efficiency becomes a critical consideration alongside data efficiency. The focus shifts towards algorithms that can efficiently leverage prior knowledge and past experiences to converge rapidly towards high-quality solutions within a shorter timeframe~\cite{coping-with-big-data-MAB-AMTEA}. The ability to adapt and transfer knowledge effectively is crucial for prompting decision-making and achieving near-optimal solutions.

Addressing the challenge of big velocity in source task-instances necessitates the development of agile and time-efficient transfer optimization algorithms. Techniques that capitalize on the temporal context of the target task and leverage sophisticated transfer learning strategies can enable algorithms to make swift and informed decisions. Integration of real-time optimization methodologies and online learning approaches can further enhance the time efficiency of transfer optimization algorithms.

\subsection{Summary}

Having thoroughly examined the characteristics discussed in the preceding sections, we propose a problem test suite comprising three practical optimization problems. Each problem is selected to represent distinct facets of complexity and diversity commonly encountered in real-life applications. 
The knapsack problem is a classical combinatorial (discrete) challenge with a focus on \textit{Big Volume} of previously solved source tasks. The planar arm problem represents a continuous optimization scenario, characterized by both \textit{Big Volume} and \textit{Big Variety} of source tasks. Minimalistic attacks encompass mixed optimization problems that exhibit characteristics of \textit{Big Volume} and \textit{Big Velocity}. The properties of the aforementioned problems can be summarized as follows:

\begin{itemize}
    \item Knapsack Problem:
    \begin{itemize}
        \item Representation: Discrete
        \item Characteristic: Big Volume
    \end{itemize}
    \item Planar Arm Problem:
    \begin{itemize}
        \item Representation: Continuous
        \item Characteristic: Big Volume and Big Variety
    \end{itemize}
    \item Minimalistic Attacks:
    \begin{itemize}
        \item Representation: Mixed (Continuous and Discrete)
        \item Characteristic: Big Volume and Big Velocity
    \end{itemize}
\end{itemize}

\begin{figure}
    \centering
    \includegraphics[width=0.50\textwidth]{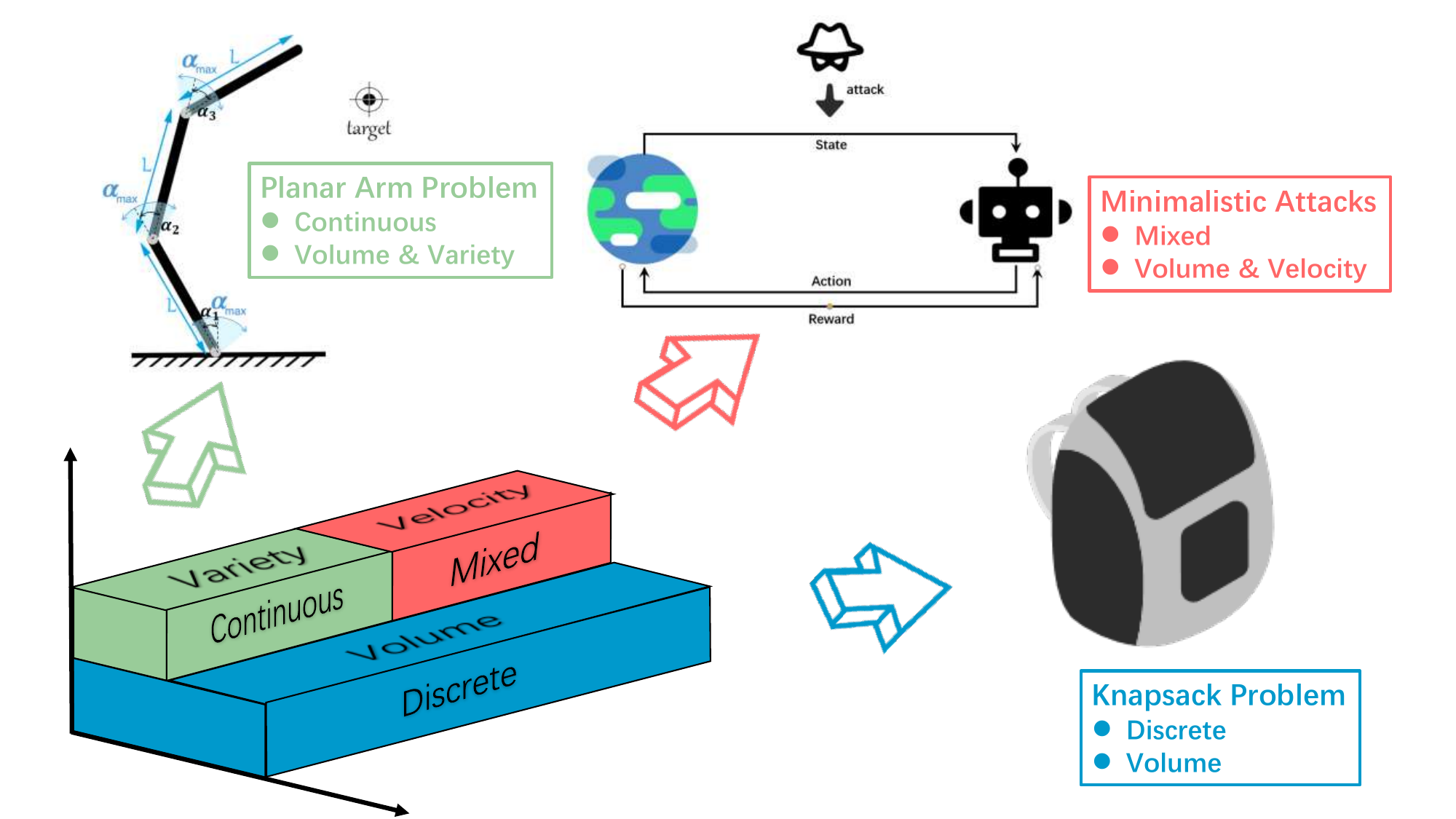}
    \caption{Categorization of our benchmarking problem suite, which comprises three representative problems, namely, the Knapsack Problem, Planar Robotic Arm Problem, and Minimalistic Attacks. These problems exemplify discrete, continuous, and mixed optimization domains, respectively. Remarkably, Knapsack Problems have only the \emph{volume} characteristic. While, Planar Robotic Arm Problems possess \emph{volume} and \emph{variety} features, and Minimalistic Attacks exhibit \textit{volume} and \emph{velocity} characteristics.}

    \label{problem-selection-figure}
\end{figure}

The visualization of the benchmark suite categories is depicted in Fig.~\ref{problem-selection-figure}. To assess the adaptability of algorithms and address the challenges posed by Big Source Task-Instances in TrEO, we examine two transfer scenarios: \emph{multi-to-one} and \emph{many-to-one.}

In the multi-to-one scenario, the process starts with active source task selection, in contrast to the one-to-one transfer scenario, where the algorithm directly transfers knowledge from a given source task without selection considerations. The multi-to-one scenario often involves a higher proportion of relevant source tasks, facilitating the algorithm's ability to identify pertinent sources effectively. Additionally, this scenario provides an opportunity to leverage knowledge from all source tasks, enabling the extraction of valuable insights while mitigating the risk of negative transfer. Conversely, the many-to-one transfer scenario involves a substantial number of source tasks, thereby challenging the scalability of algorithms. This scenario also tests the agility of algorithms when faced with variations in the sparsity of relevant source tasks for the target task of interest. In what follows, we present detailed descriptions of the proposed optimization problems, capturing diverse complexities encountered in real-world applications.

\section{Problems Description}
\label{probdescription}

In this section, we present a concise introduction to the three practical problems included in our benchmarking problem suite: the 0/1 knapsack problem, the planar robotic arm problem, and minimalistic attacks. Additionally, we outline the methodology used to measure source-target correlation/similarity within these problems.

\subsection{Constructing 0/1 Knapsack Problem Benchmarks}
The knapsack problem is one of the most studied discrete NP-hard problems~\cite{knapsack-problems:}. A formal definition of the 0/1 knapsack problem is given below:

\begin{equation}\label{KP}
\begin{aligned}
     &\mathop{\max}\sum_{i=1}^n v_ix_i, \\
     s.t. \sum_{i=1}^n &w_ix_i \leq C \  and \  x_i\in \{0, 1\},
\end{aligned}
\end{equation}
where $n$ indicates the number of indivisible items, $v_i$ represents the value of the $i^{th}$ item, and $w_i$ denotes the weight of the $i^{th}$ item. The binary variable $x_i$ is used to indicate whether the $i^{th}$ item is selected ($x_i=1$) or not ($x_i=0$). When employing EAs to solve the knapsack problem, there is a possibility of violating the capacity constraint for certain individuals. In such cases, we employ Dantzig's greedy approximation algorithm to ensure compliance with the capacity constraint~\cite{10.1007/3-540-58495-1_14}.

Practical problems with similarities to knapsack problems are widespread, and researchers are actively seeking more efficient solutions that require less time and computational resources. For example, in a cloud service that recommends packages to clients,  the process of finding customized packages to recommend to a new target client can be cast as the NP-hard 0/1 knapsack problem~\cite{sTrEO}. However, many current algorithms still approach these problems from scratch, leading to significant time and computational demands. This is where transfer optimization algorithms come into play, offering valuable benefits as problems rarely exist in isolation and often share common mathematical abstractions. By leveraging knowledge from previously solved tasks, transfer optimization can greatly enhance the optimization process and alleviate the burden of starting the search from scratch each time.

In this study, we generate synthetic instances for the knapsack problem (KP) with attributes similar to those found in~\cite{10.1007/3-540-58495-1_14, curbing-AMTEA, coping-with-big-data-MAB-AMTEA}. We categorize the instances based on the relationships between $w_i$ and $v_i$ and randomly generate three instances as follows: (1) \textit{uncorrelated} (uc), where both $w_i$ and $v_i$ are uniformly generated random real numbers within the range [1, 10]; (2) \textit{weakly correlated} (wc), where $w_i$ is a uniformly generated random real number within [1, 10], and $v_i$ is obtained by adding $w_i$ to another uniformly generated random real number within the range [-5, 5] (if, for any $i$, $v_i < 0$, we discard the sample and repeat the process until $v_i \ge 0$); and (3) \textit{strongly correlated} (sc), where $w_i$ is still a uniformly generated random real number within [1, 10], and $v_i$ is defined as $w_i + 5$. Additionally, we define two types of knapsacks based on their capacity: (1) \textit{restrictive capacity} (rk), where $C = 20$ and only a small number of items can be selected; and (2) \textit{average capacity} (ak), where $C = 0.5\sum_{i=1}^n w_i$ and the number of items is larger. 

Based on the categorization above, we can build six different types of KP tasks (i.e., KP\_uc\_rk, KP\_wc\_rk, KP\_sc\_rk, KP\_uc\_ak, KP\_wc\_ak, and KP\_sc\_ak). Note two tasks are considered to be more closely related when they both have an average capacity. This often manifests in a significant degree of similarity between them as a relatively large number of items need to be selected for both tasks and hence their optimum solutions could depict a substantial overlap~\cite{coping-with-big-data-MAB-AMTEA}.

\subsection{Planar Robotic Arm Problem}\label{arm_description_correlation}

For this study, we selected a 2D robotic arm problem inspired by~\cite{quality-diversity}. This problem entails a robotic arm comprised of $d$ joints, where each joint possesses an identical length $L$ and the ability to rotate up to a maximum angle $\alpha_{max}$ (encoded in (0,1]). The objective is to determine suitable angles for each joint, within the specified boundaries, in order to position the tip of the robotic arm at a given target point. This scenario resembles a human arm reaching out to grasp an object.

To solve the problem, we aim to determine the optimal angles $\alpha = (\alpha_1, \alpha_2, ..., \alpha_d)$ for each joint such that the position of the robotic arm's tip is as close as possible to a predefined target in the two-dimensional plane. The dimensionality of the problem denoted as $d$, corresponds to the number of joints or links in the robotic arm. The number of joints thus introduces variety by instilling heterogeneity in the search spaces of different tasks.

We define the task $\mathcal{T}_{L,\alpha_{max}}$ by two parameters: the length of the links, denoted as $L$ (assumed to be the same for all links), and the maximum angle $\alpha_{max}$ for each joint (assuming equal limits for all joints for simplicity). The objective function, $f(\alpha, \mathcal{T}_{L,\alpha_{max}})$, evaluates a solution $\alpha$ in the context of the task $\mathcal{T}_{L,\alpha_{max}}$. It computes the Euclidean distance between the tip position, denoted as $p_d$, and the target position denoted as $T$. The recursive calculation of $p_d$ is as follows:

\begin{equation}\label{planar_arm_initial_M}
     M_0\ =\ I ,
\end{equation}
\begin{equation}\label{planar_arm_M_iteration}
    M_i\ =\ M_{i-1}\ \cdot\  
  \begin{pmatrix}
    \cos{\alpha_i^\prime} & -\sin{\alpha_i^\prime} & 0 & L^\prime\\
    sin{\alpha_i^\prime} & \cos{\alpha_i^\prime} & 0 & 0\\
    0 & 0 & 1 & 0\\
    0 & 0 & 0 & 1
  \end{pmatrix},
\end{equation}
\begin{equation}\label{planar_arm_p_iteration}
    p_i\ =M_i\ \cdot\ {(0,\ 0,\ 0,\ 1)}^T ,
\end{equation}
where, $\alpha_i = 2\pi \cdot \alpha_{max} \cdot (\alpha_i - 0.5),\ \forall i\in \{1, ..., d\}$ and $L^\prime\ =\ L/d$. Under these settings, we can describe the objective function as follows:
\begin{equation}\label{planar_arm_objective}
     f(\alpha,\mathcal{T}_{L,\alpha_{max}}\ )\ =\ -\Vert p_d - T \Vert .
\end{equation}
Here, we define the target position $T$ as the point (1, 1) in the two-dimensional plane. Moreover, we consider this as a maximization problem, aiming to maximize the negative distance between the tip position $p_d$ and the target position $T$ in equation~(\ref{planar_arm_objective}). 

\begin{figure}
    \centering
    \subfigure[]{
    \begin{minipage}{7.5cm}
    \centering
    \includegraphics[width=7.5cm]{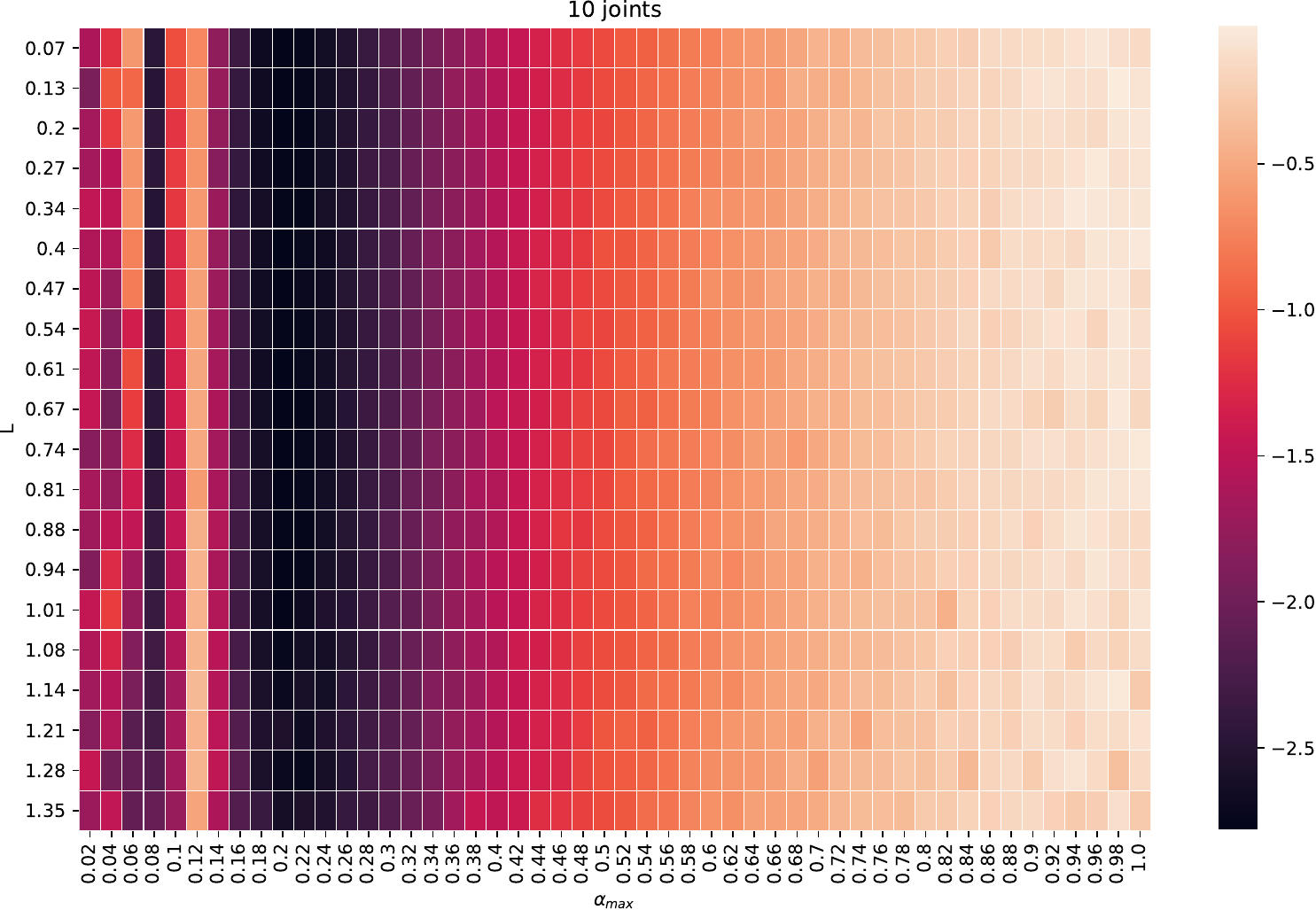}
    \end{minipage}
    }
    \\
    \subfigure[]{
    \begin{minipage}{7.5cm}
    \centering
    \includegraphics[width=7.5cm]{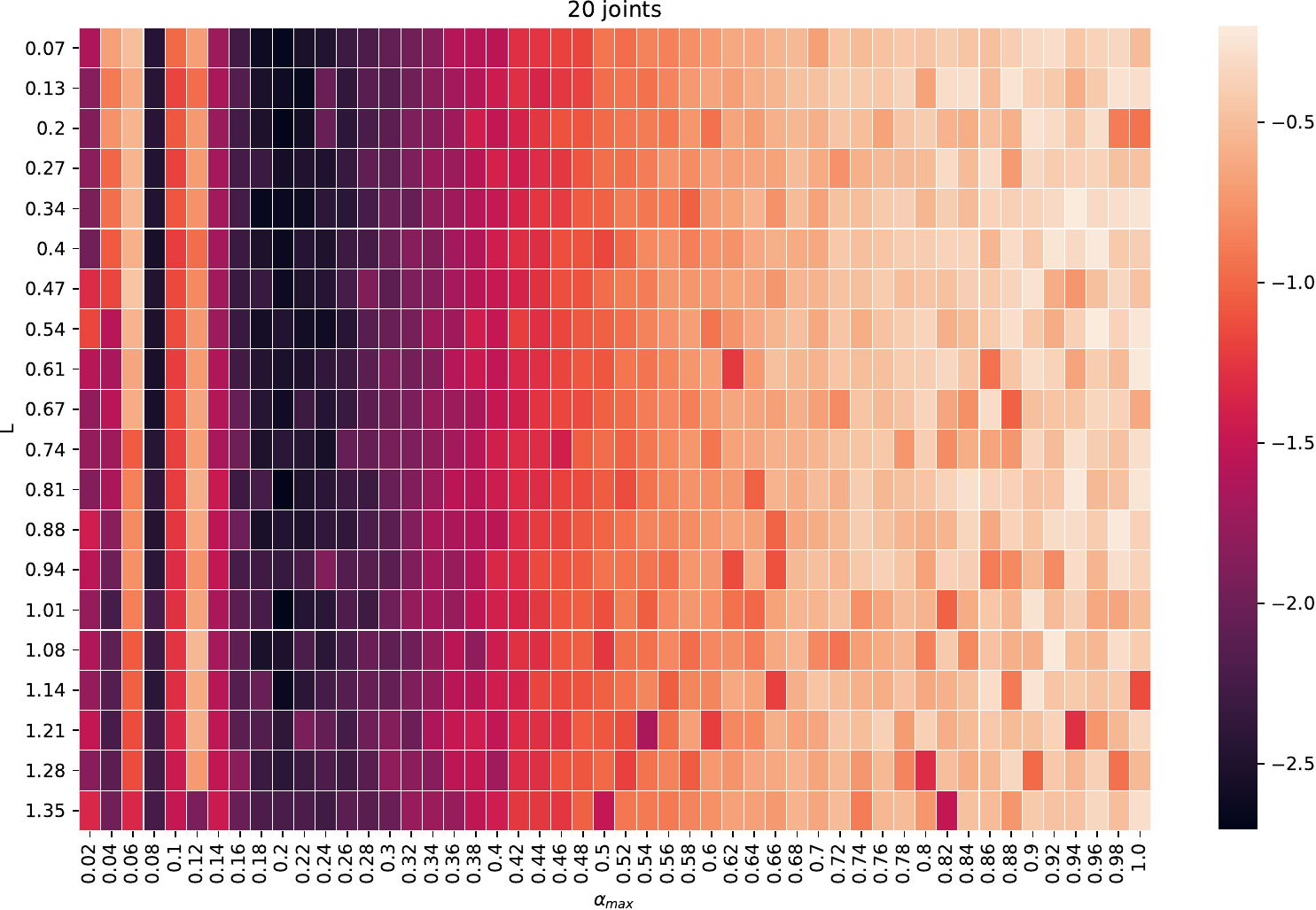}
    \end{minipage}
    }
    \caption{Heatmaps of source-target relatedness for diverse combinations of $L$ and $\alpha_{max}$, ranging between (0,$\sqrt{2}$) and (0, 1], respectively, for (a) 10 joints and (b) 20 joints. According to the sidebar on the right which marks the fitness range, the hotter (brighter) cells indicate more relatedness to the target (as their fitness values are greater) whereas the colder (darker) ones show less or no relatedness.}
    \label{heatmap}
\end{figure}

After clarifying the definition of the planar robotic arm problem, understanding the similarity/correlation between the tasks is necessary for designing benchmarking problems in the transfer optimization domain. In the present study, we consider two cases involving 10 and 20 joints. The target task is defined as $\mathcal{T}_{\sqrt{2},1}$ for both scenarios.
We generate numerous source instances with different lengths $L$ in the range (0, $\sqrt{2}$) and angles $\alpha_{max}$ in the range (0, 1). The similarity estimation process between source and target tasks is performed by the following steps. First, we optimize all the source tasks using a continuous canonical genetic algorithm (CGA)~\cite{cga}. Then, after constructing probabilistic models~\cite{mvarnorm} of the source tasks, a number of solutions are sampled from each source task model and evaluated using the objective function of the target task. The objective values form the cells of the heat maps, as shown in Fig.~\ref{heatmap}, which provide valuable insights into the correlation between the source and target tasks. Note that the dimensionalities of the source and target tasks are kept the same to ensure the reliability of the similarity estimation.

In the heat maps, cells with higher fitness values appear hotter (brighter), indicating a stronger correlation with the target task. Conversely, colder (darker) cells represent weaker or no correlation due to the useless or negative knowledge transfer. Essentially, the degree of correlation between the source tasks and the target task is determined by the effectiveness of the transfer process. By analyzing this heat map, we gain a clear understanding of the similarity between the source task constructed with specific $\alpha_{max}$ and $L$ and the target task. This valuable insight not only streamlines the creation of the benchmark for the planar robotic arm problems but also lays the groundwork for its future development.

\subsection{Minimalistic Attacks}
\label{sec:Minimalistic Attacks}

The rapid advancements in artificial intelligence (AI) have led to significant breakthroughs, but they have also brought forth new challenges, particularly in AI security. One such challenge is exemplified in~\cite{minimalistic-attack}, where the vulnerability of neural network-based reinforcement learning policies is exposed. Moreover, recent research showcases the transferability of perturbations applied to similar frames, enabling attacks on specific frames~\cite{frame-correlation}. 
In light of the above, we have incorporated minimalistic attacks as the third test problem in our benchmarking suite. This problem provides a valuable platform for evaluating the algorithms' ability to deceive systems and create adversarial attacks in realistic AI scenarios.

\begin{figure}
    \centering
    \subfigcapskip=5pt 
    \subfigure[Original State]{
    \begin{minipage}{4cm}
    \centering
    \includegraphics[width=4cm]{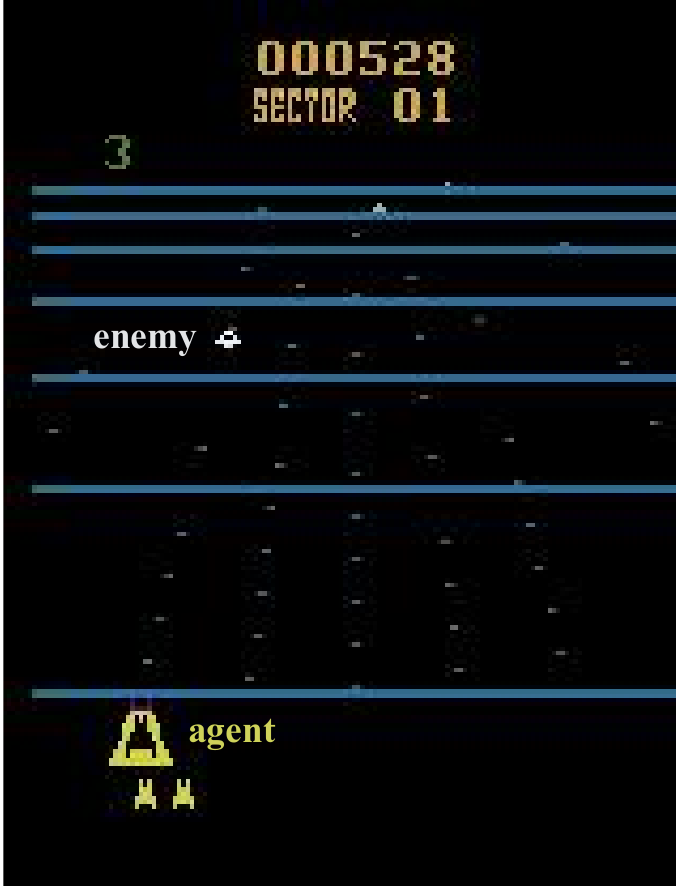}
    \end{minipage}
    }
    \subfigure[Attacked State]{
    \begin{minipage}{4cm}
    \centering
    \includegraphics[width=4cm]{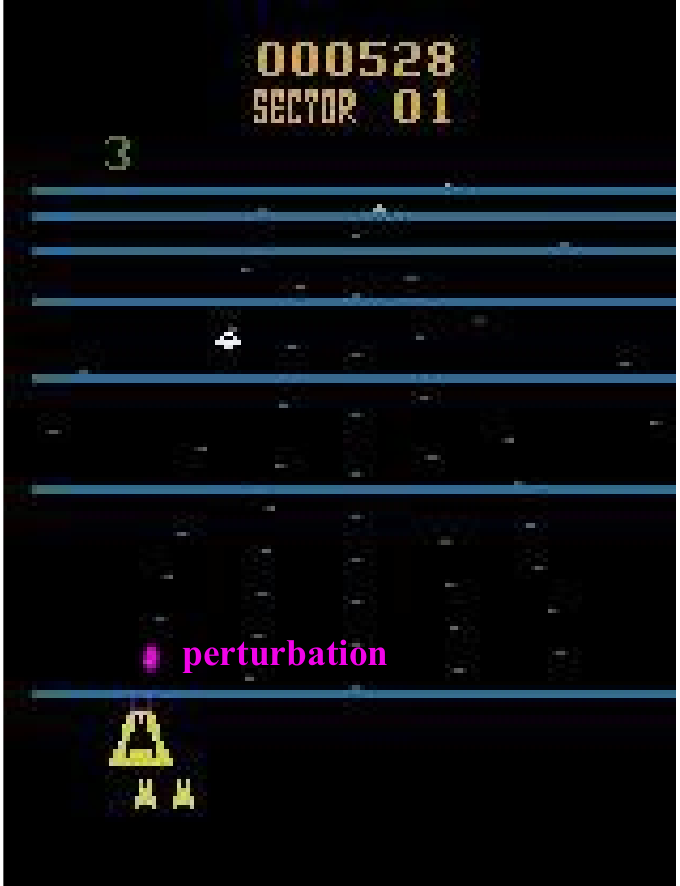}
    \end{minipage}
    }
    \caption{Minimalistic Attacks: In the original BeamRider keyframe (a), an enemy is directly in front of the agent. At this point, the trained agent fires, resulting in an increase in the game's score.  However, if the keyframe is disturbed due to an attack (b), causing the agent to mistakenly perceive a bullet, the trained agent becomes more inclined to move left or right to avoid it.  Consequently, this alteration in behavior ultimately leads to a successful attack as the agent's original actions are changed.}
    \label{minimalistic-attacks}
\end{figure}

Minimalistic attacks represent a type of reinforcement learning (RL) adversarial attack problem that encompasses three key settings~\cite{minimalistic-attack}: (1) black-box policy access; (2) fractional-state adversary, and (3) tactically-chanced attack.
The main objective of these attacks is to deceive RL policies by perturbing a specific number of pixels in selected keyframes, assuming a black-box setting (see Fig.~\ref{minimalistic-attacks}). 

In an RL example, the agent takes action $a_t$ based on the state $s_t$ and receives a reward $r_t$ from the environment at time step $t$. Assuming a finite set of $n$ available actions $a_t^1, a_t^2, ..., a_t^n$, the action probability distribution $\pi(\cdot|s_t)$ over those $n$ actions can be described as: 
\begin{equation}\label{actionDistribution}
    \begin{aligned}
        \pi(\cdot|s_t) &= [p(a_t^1), p(a_t^2), ..., p(a_t^n)], \\ 
        &s.t. \sum_{j=1}^n p(a_t^j) = 1. 
    \end{aligned}
\end{equation}

Herein, $p(a_t^j)$ represents the probability that the agent chooses action $a_t^j$. As expected, the agent selects the action $o = \arg\max_j p(a_t^j)$. With this, the goal of minimalistic attacks is to maximize the discrepancy between action distributions before and after the attack, which can be formulated as follows:

\begin{equation}\label{attack_formulation}
   \ \mathop{max}\limits_{\delta_t}\ \mathop{max}\limits_{e \ne o}\ \pi (\cdot | s_t + \delta_t)_e - \pi (\cdot | s_t + \delta_t)_o,
\end{equation}
where $\delta_t$ represents the perturbation to be added to the original state $s_t$ at time step $t$. Further, $o$ and $e$ represent the action taken by the trained agent before and after the attack and $\pi (\cdot | s_t )_e$ represents the probability that the agent chooses action $j$ under the guidance of $\pi(\cdot|s_t)$. In the context of minimalistic attacks, $\delta_t$ is limited to perturb only a small fraction of the input state. 

Remarkably, a successful attack is achieved when the fitness value in equation~(\ref{attack_formulation}) is greater than 0, indicating that the agent has been deceived and takes a different action.
Under such circumstances, the time for the algorithm to compute an attack is relatively short due to the transient nature of the frames. 
Recently, researchers introduced a novel approach to accelerate the computation of perturbation $\delta_t$ by exploiting the correlation between similar frames~\cite{frame-correlation}. This technique, referred to as Init-GA, involves incorporating the best-performing individual (with the highest fitness value) from the final population of an already optimized attack into the initial population of the target attack. This method effectively utilizes correlations between the source and target tasks, resulting in more efficient optimization.

Based on these insights, we classify the source tasks into two groups: those that strongly correlate with the target task and those that weakly correlate. This is determined by the observed transfer outcomes during the optimization process.
To elaborate, our approach involves two steps: (1) optimizing each source task using CGA, and (2) sequentially transferring an individual (or candidate solution) from the final population of optimized source tasks to the initial population of the target task. This sequential transfer process repeats until each individual in the source task's final population has been transferred to the target task. 
For each source task, we conduct an equal number of independent experiments as its population size. It's worth noting that only a single individual is transferred from the source task at a time. An individual is considered positive if it contributes to the successful completion of an attack within a reduced number of generations during the optimization of the target task. Moreover, we define a source task as related to the target task when the majority (i.e., at least 60\%) of its population consists of positive individuals. To ensure statistical significance and reduce the influence of random noise, we repeat this process 30 times.

\section{problem specification and baseline results}
\label{baseline}

This section provides a detailed discussion of our benchmark problem settings, thoroughly testing the algorithms' capabilities in solving the challenges of \textit{Big Volume}, \textit{Big Variety}, and \textit{Big Velocity}. Besides, we provide comprehensive results and analysis of state-of-the-art algorithms for comparison and evaluation.
 
\subsection {Big Volume}

\emph{Multi-to-one Problem:}
In this scenario, we consider three different configurations for defining source and target tasks (see Table~\ref{multi-KP-configurations}). Each configuration includes four types of KPs as source tasks. Note that only the last source task with average knapsack capacity (labeled as KP\_sc\_ak (task4), KP\_uc\_ak (task4), and KP\_wc\_ak (task4)) is strongly related to their corresponding target task, while the others are weakly related. 

\begin{table}[htbp]
    \newcommand{\tabincell}[2]{\begin{tabular}{@{}#1@{}}#2\end{tabular}}
    \centering
    \caption{Configuration details of multi-to-one scenario for the 0/1 knapsack problem.}
    \setlength{\tabcolsep}{1.5mm}{
        \small
        \begin{tabular}{ccc}
            \toprule
            Configuration & Source Tasks & Target Task \\
            \midrule
            A & \tabincell{c}{KP\_uc\_rk(task1), KP\_sc\_rk(task2), \\ KP\_wc\_rk(task3), KP\_sc\_ak(task4)} & KP\_uc\_ak \\
            B & \tabincell{c}{KP\_uc\_rk(task1), KP\_sc\_rk(task2), \\ KP\_wc\_rk(task3), KP\_uc\_ak(task4)} & KP\_wc\_ak \\
            C & \tabincell{c}{KP\_uc\_rk(task1), KP\_sc\_rk(task2), \\ KP\_wc\_rk(task3), KP\_wc\_ak(task4)} & KP\_sc\_ak \\
            \bottomrule
        \end{tabular}
        }
    \label{multi-KP-configurations}
\end{table}

\emph{Many-to-one Problem:}
In this scenario, the number of source tasks is 1000 in alignment with the big volume condition. The type of source tasks include {KP\_uc\_rk, KP\_sc\_rk,  KP\_wc\_rk, KP\_sc\_ak} and the target belongs to KP\_uc\_ak. We consider four different settings where the related tasks (i.e., KP\_sc\_ak) account for 22\%(220), 16\%(160), 10\%(100), and 4\%(40) of the total of 1000 source tasks, respectively. 
This can assess the algorithm's performance when dealing with sparse relationships between the source and target tasks. As the ratio of related source tasks decreases, the algorithm's ability to extract useful knowledge from a large \emph{volume} of previously optimized tasks is demanded.

\emph{Experimental Settings:}
For each knapsack problem, the dimensionality is set to 2000, meaning there are 2000 items available for selection. 
While there are a plethora of existing methods for addressing transfer optimization problems, applying them directly in the context of discrete 0/1  knapsack problems proves to be challenging. 
In this work, we choose to utilize the following four methods: (\textit{i}) Canonical Genetic Algorithm~\cite{cga} (CGA, a classical genetic algorithm with no transfer), (\textit{ii}) Evolutionary Knowledge Transfer~\cite{EKT} (EKT, a simple transfer through population seeding/case-injection), (\textit{iii}) Adaptive Model-based Transfer Evolutionary Algorithm~\cite{curbing-AMTEA} (AMTEA, a model-based transfer with stacked density estimation), and (\textit{iv}) scalable Transfer Evolutionary Optimization~\cite{sTrEO} (sTrEO, a model-based transfer with two co-evolving species for joint evolution). In our experiments, we begin by optimizing the source tasks of each configuration using the binary CGA~\cite{cga}. The populations of the first and final generations are saved for aiding target task optimization later on. This process incorporates a stopping condition to ensure the diversity of the population. The optimization of the knapsack problem is carried out under the following settings:

\begin{enumerate}
    \item Representation: Binary coded.
    \item Repetition: 30.
    \item Population size: 50.
    \item Maximum function evaluations: 5000.
    \item Evolutionary operators:
    \begin{enumerate}
        \item Uniform crossover~\cite{uniform-crossover} with probability $p_c$=1.
        \item Bit-flip mutation with probability $p_m=1/d$, where $d$ is the dimensionality of the target optimization problem.
    \end{enumerate}
    \item For model-based transfer algorithm(s):
    \begin{enumerate}
        \item Probabilistic model: Univariate marginal frequency (factored
Bernoulli distribution)~\cite{umd}.
        \item Transfer interval: 2.
    \end{enumerate}
\end{enumerate}

All source tasks and the target task employ CGA of the same genetic operators (as listed above) as the basic solvers. Moreover, the settings (i.e., hyper-parameters) of AMTEA and sTrEO in the experiments are kept consistent with previous studies~\cite{curbing-AMTEA, sTrEO}.

\emph{Performance Metrics:}\label{knapsack-metrics} In knapsack problems, we consider the averaged objective value, performance score, and wall clock time as the metrics to measure the performance of all algorithms. Suppose there are $N$ optimization algorithms applied to this problem, and each algorithm runs for $L$ repetitions. The final objective value obtained by the $i^{th}$ algorithm in the $l^{th}$ repetition for the optimization problem is denoted as $y_{il}$, where $i \in {1, 2, ..., N}$ and $l \in {1, 2, ..., L}$. The three metrics of the $i^{th}$ algorithm can be given as:

\begin{itemize}
    \item Averaged Objective Value: 
    The averaged objective value means the average fitness of the population and can be computed as $\frac{1}{L}\sum_{l=1}^L y_{i,l}$. A higher average objective value is better for optimizing a maximization problem.
    \item Performance Score: 
    To calculate the performance score for each algorithm, we compute the mean ($\mu$) and standard deviation ($\sigma$) of the objective values obtained by all $N$ algorithms over the $L$ repetitions. Upon which, the performance score for the $i^{th}$ algorithm is calculated as follows:
    
    \begin{equation}\label{score}
    score_i = \frac{1}{L}\sum_{l=1}^L \frac{y_{il} - \mu}{\sigma}.
    \end{equation}
     The highest performance score refers to its corresponding algorithm outperforming the others.
    \item Wall Clock Time: 
    The wall clock time represents time cost during the optimization process. A smaller wall clock time implies an algorithm can optimize the problem at a lower time cost.
\end{itemize}

\emph{Comparison Results and Analysis:} 

(1) \textbf{Data Efficiency}: Tables~\ref{multi-knapsack-results} and~\ref{many-knapsack-results} summarize the averaged objective values and performance scores of the CGA, EKT, AMTEA, and sTrEO on completing 5000 function evaluations. The best values are shown in {\bf boldface}. 
The results show that there is a small difference between the averaged objective values obtained by the four above optimization algorithms. 
Overall, in the multi-to-one transfer scenario, AMTEA is slightly superior to the others. However, in the many-to-one transfer scenario, sTrEO obtained the most competitive performance among all methods.

\begin{table*}[!htbp]
    \renewcommand\arraystretch{1.5}
    \centering
    \caption{Averaged objective values and performance scores of CGA, EKT, AMTEA, and sTrEO for three configurations of multi-to-one scenario in the 0/1 Knapsack Problem. The best results are shown in \textbf{boldface}.}
    \setlength{\tabcolsep}{5mm}{
        \scalebox{0.75}{
        \begin{tabular}{ccccccccccc} 
            \toprule 
            \multicolumn{3}{c}{\multirow{2}*{Configuration}} & \multicolumn{4}{c}{Averaged Objective Value} &
            \multicolumn{4}{c}{Performance Score}\\
            \multicolumn{3}{c}{} & CGA & EKT & AMTEA & sTrEO & CGA & EKT & AMTEA & sTrEO  \\ 
            \hline 
            \multicolumn{3}{c}{A} & 7998.3247 & 7985.1407 &  \textbf{8156.0013} & 8131.5400 & -0.8353 & -0.9939 & \textbf{1.0617} & 0.7674\\ 
            \multicolumn{3}{c}{B} & 7704.2353 & 7698.5887 &  \textbf{7767.6047} & 7750.6513 & -0.7186 & -0.8745 &  \textbf{1.0306} & 0.5626\\
            \multicolumn{3}{c}{C} & \textbf{11910.6927} & 11885.8580 &  11895.2360 & 11759.3393 & \textbf{0.6655} & 0.3205 &  0.4508 & -1.4368\\
            \bottomrule 
        \end{tabular}
        }
    }
    \label{multi-knapsack-results}
\end{table*}

\begin{table*}[!htbp] 
    \renewcommand\arraystretch{1.5}
    \centering
    \caption{Averaged objective values and performance scores of CGA, EKT, AMTEA, and sTrEO for four configurations of many-to-one scenarios in the 0/1 Knapsack Problem. The best results are shown in \textbf{boldface}.}
    \setlength{\tabcolsep}{6.2mm}{
        \scalebox{0.75}{
        \begin{tabular}{ccccccccccc} 
            \toprule 
            \multicolumn{3}{c}{\multirow{2}*{Ratio}} & \multicolumn{4}{c}{Averaged Objective Value} &
            \multicolumn{4}{c}{Performance Score}\\
            \multicolumn{3}{c}{} & CGA & EKT & AMTEA & sTrEO & CGA & EKT & AMTEA & sTrEO  \\  
            \hline 
            \multicolumn{3}{c}{0.22} & 7994.7720 & 8000.9487 & 8322.0060 & \textbf{8419.5040} &  -0.9833 & -0.9513 & 0.7144 & \textbf{1.2202}\\    
            \multicolumn{3}{c}{0.16} & 7987.0113 & 7998.7113 & 8316.3527 & \textbf{8421.2513} & -0.9950 & -0.9349 & 0.6957 & \textbf{1.2342}\\    
            \multicolumn{3}{c}{0.10} & 8005.4233 & 8000.4120 & 8304.0240 & \textbf{8401.4547} & -0.9472 & -0.9747 & 0.6933 & \textbf{1.2286}\\
            \multicolumn{3}{c}{0.04} & 8005.8573 & 8008.0007 & 8269.9033 & \textbf{8370.6260} & -0.9527 & -0.9398 & 0.6421 & \textbf{1.2504}\\
            \bottomrule 
        \end{tabular}
        }
    }
    \label{many-knapsack-results}
\end{table*}

\begin{figure*}[htbp]
    \centering
    \subfigure[]{
    \begin{minipage}{5cm}
    \centering
    \includegraphics[width=5cm]{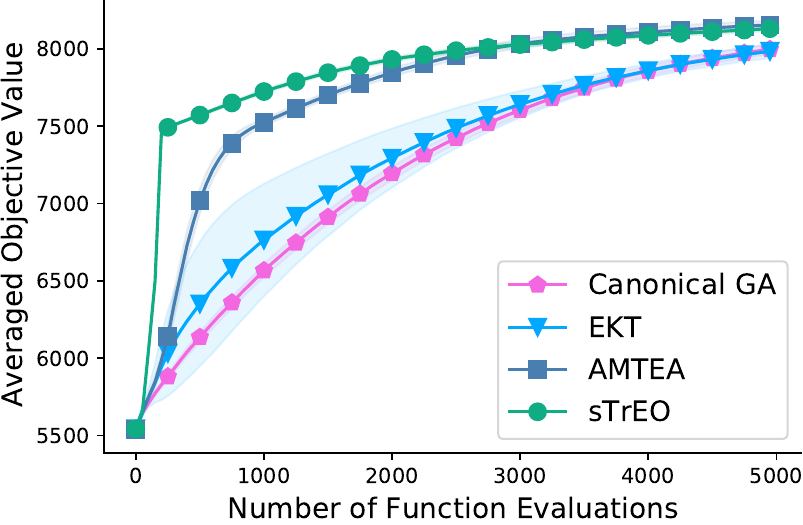}
    \end{minipage}
    }
    \subfigure[]{
    \begin{minipage}{5cm}
    \centering
    \includegraphics[width=5cm]{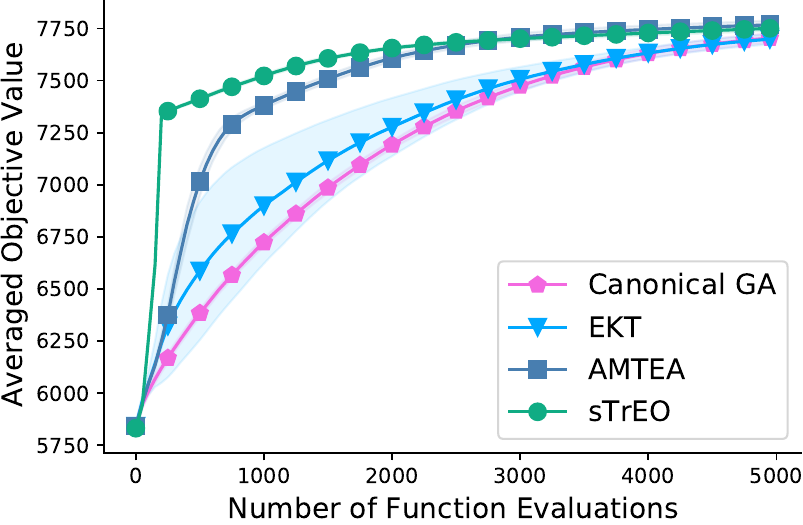}
    \end{minipage}
    }
    \subfigure[]{
    \begin{minipage}{5cm}
    \centering
    \includegraphics[width=5cm]{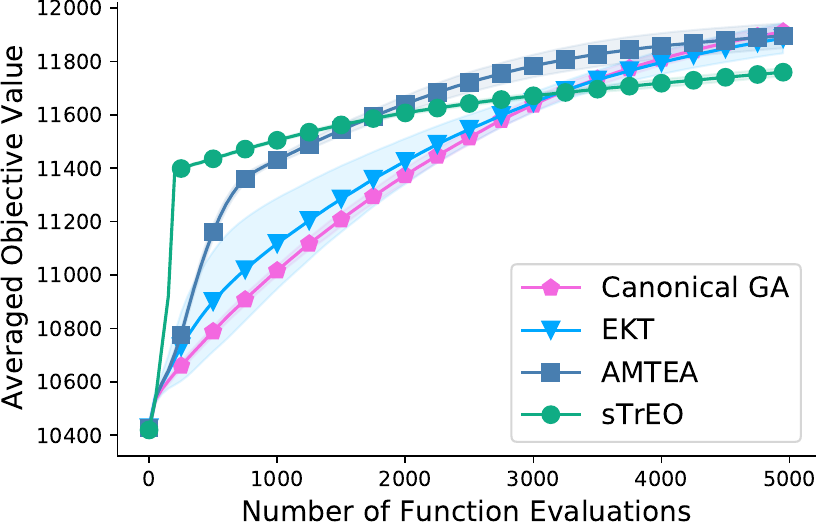}
    \end{minipage}
    }
    \caption{Convergence trends for (a) configuration A, (b) configuration B, and (c) configuration C of multi-to-one scenario in the 0/1 Knapsack Problem with 5000 function evaluations.}
    \label{convergence-trends-for-multi-knapsack}
\end{figure*}

\begin{figure*}[htbp]
    \centering
    \subfigure[]{
    \begin{minipage}{4cm}
    \centering
    \includegraphics[width=4cm]{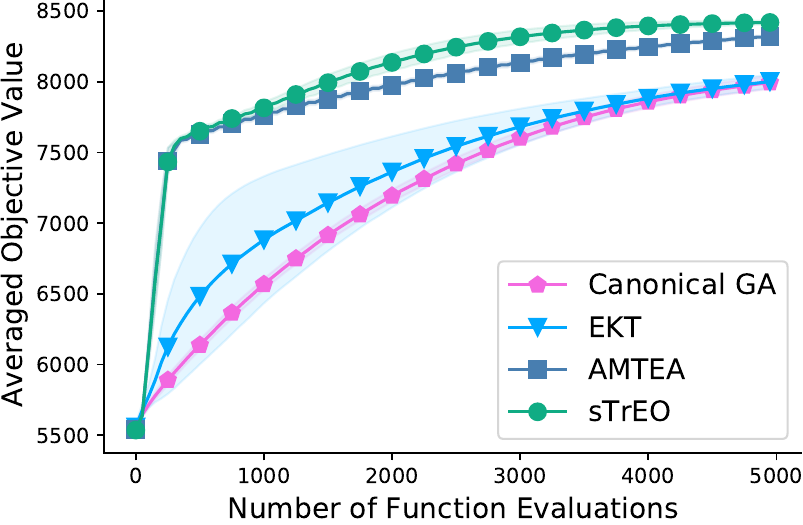}
    \end{minipage}
    }
    \subfigure[]{
    \begin{minipage}{4cm}
    \centering
    \includegraphics[width=4cm]{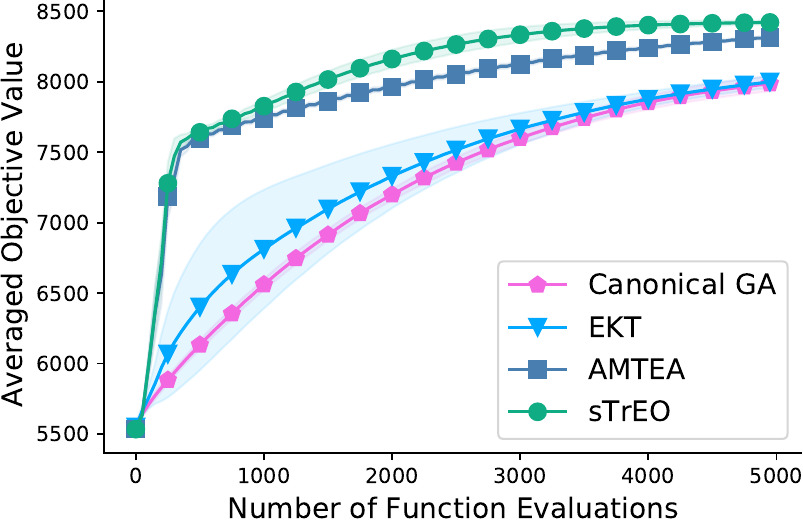}
    \end{minipage}
    }
    \subfigure[]{
    \begin{minipage}{4cm}
    \centering
    \includegraphics[width=4cm]{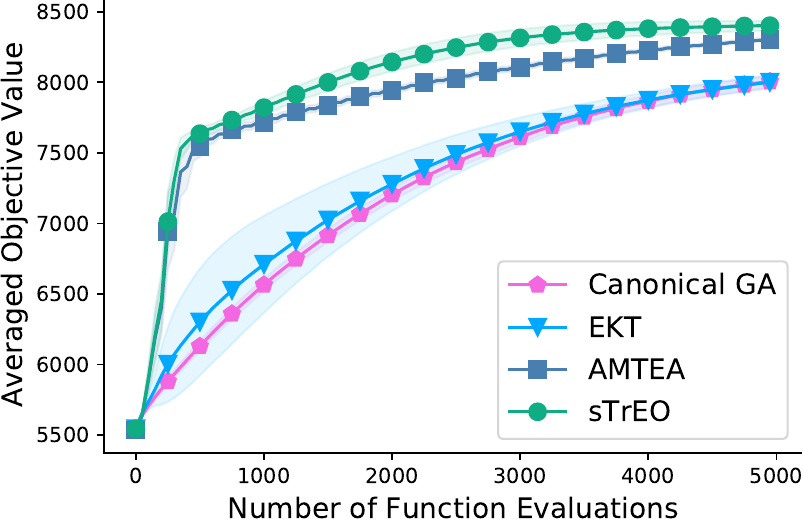}
    \end{minipage}
    }
    \subfigure[]{
    \begin{minipage}{4cm}
    \centering
    \includegraphics[width=4cm]{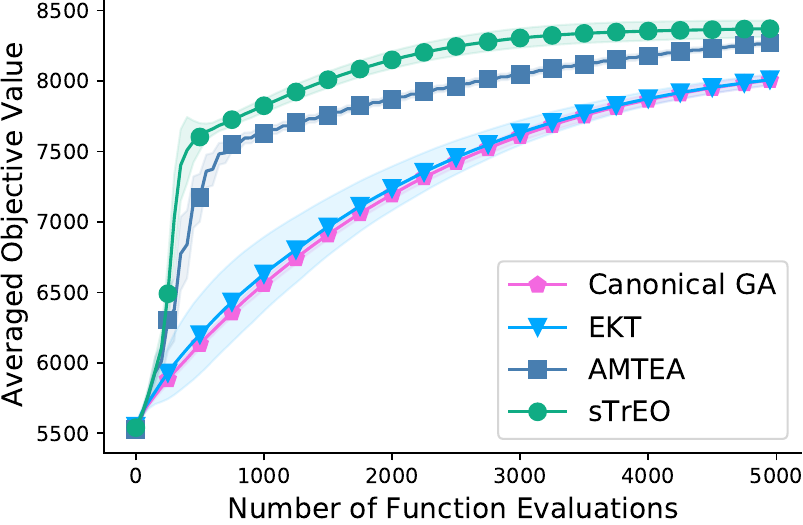}
    \end{minipage}
    }
    \caption{Convergence trends with related source tasks account for (a) 0.22, (b) 0.16, (c) 0.10, and (d) 0.04 ratios of many-to-one scenario in the 0/1 Knapsack Problem with 5000 function evaluations.}
    \label{convergence-trends-for-many-knapsack}
\end{figure*}

Fig.~\ref{convergence-trends-for-multi-knapsack} and Fig.~\ref{convergence-trends-for-many-knapsack} show the convergence trends of the averaged objective values obtained by four algorithms in the multi-to-one and many-to-one scenarios for 0/1 knapsack problems, respectively. It can be seen that CGA obtained comparatively worse convergence performance among all methods as it does not leverage the knowledge from source tasks. AMTEA and sTrEO show a faster convergence than EKT. This is because AMTEA and sTrEO can extract useful knowledge from previously solved tasks continuously, leading to more efficient optimization than EKT with only a simple transfer of solution injection at the population initialization stage. As the ratio of closely relevant source tasks decreases in the many-to-one transfer scenario, AMTEA and sTrEO still obtain a promising optimization performance, hence demonstrating their superiority in dealing with the sparsity of the ``Big Volume'' of source task instances.

\begin{figure*}[htbp]
    \centering
    \subfigure[]{
    \begin{minipage}{5cm}
    \centering
    \includegraphics[width=5cm]{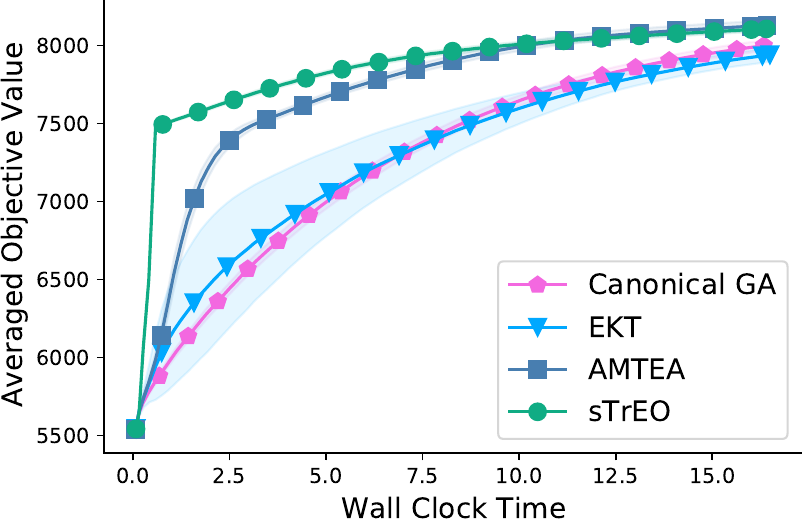}
    \end{minipage}
    }
    \subfigure[]{
    \begin{minipage}{5cm}
    \centering
    \includegraphics[width=5cm]{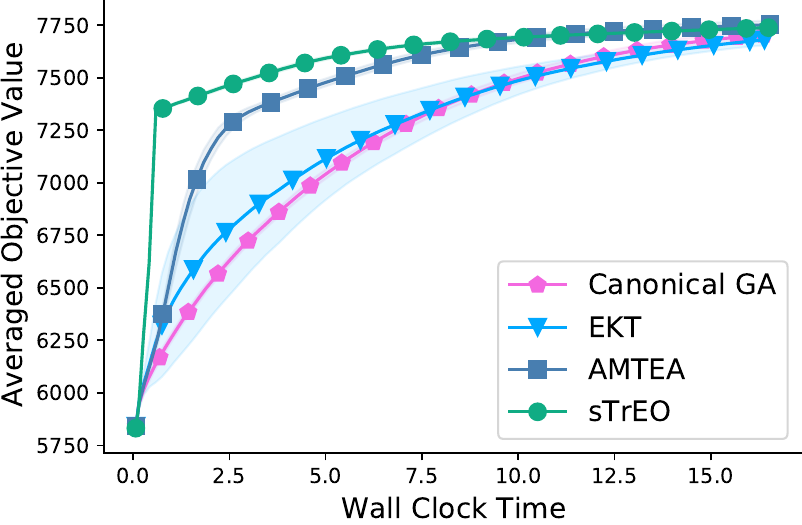}
    \end{minipage}
    }
    \subfigure[]{
    \begin{minipage}{5cm}
    \centering
    \includegraphics[width=5cm]{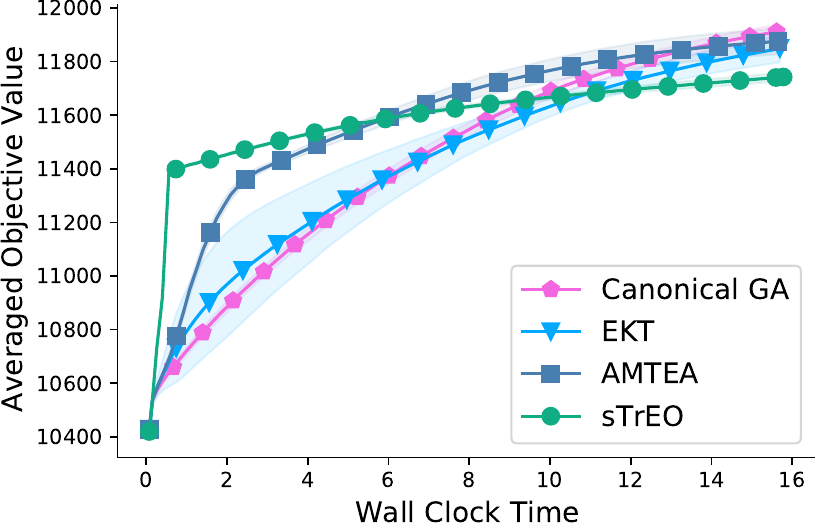}
    \end{minipage}
    }
    \caption{Convergence efficiency in terms of wall clock time in seconds for (a) configuration A, (b) configuration B, and (c) configuration C of multi-to-one scenario in the 0/1 Knapsack Problem.}
    \label{convergence-times-for-multi-knapsack}
\end{figure*}

\begin{figure*}[htbp]
    \centering
    \subfigure[]{
    \begin{minipage}{4cm}
    \centering
    \includegraphics[width=4cm]{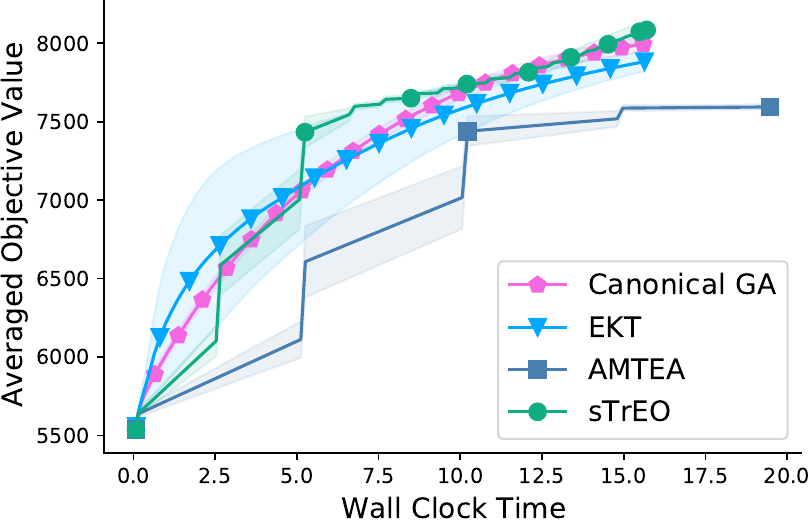}
    \end{minipage}
    }
    \subfigure[]{
    \begin{minipage}{4cm}
    \centering
    \includegraphics[width=4cm]{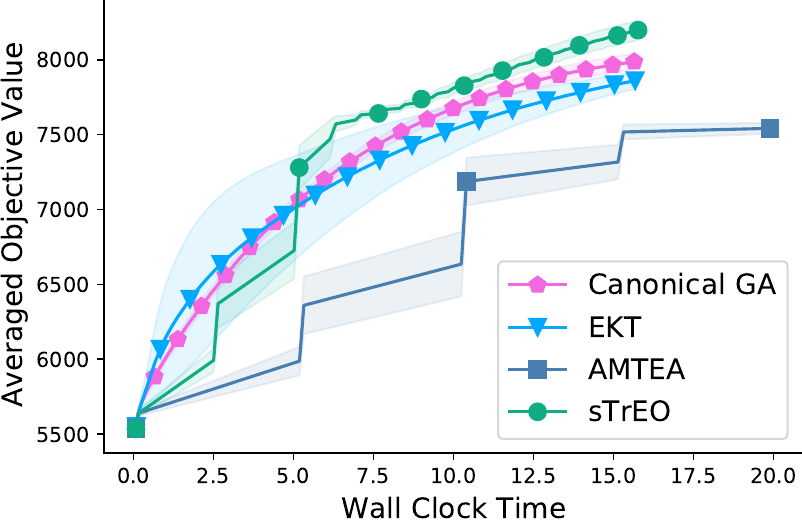}
    \end{minipage}
    }
    \subfigure[]{
    \begin{minipage}{4cm}
    \centering
    \includegraphics[width=4cm]{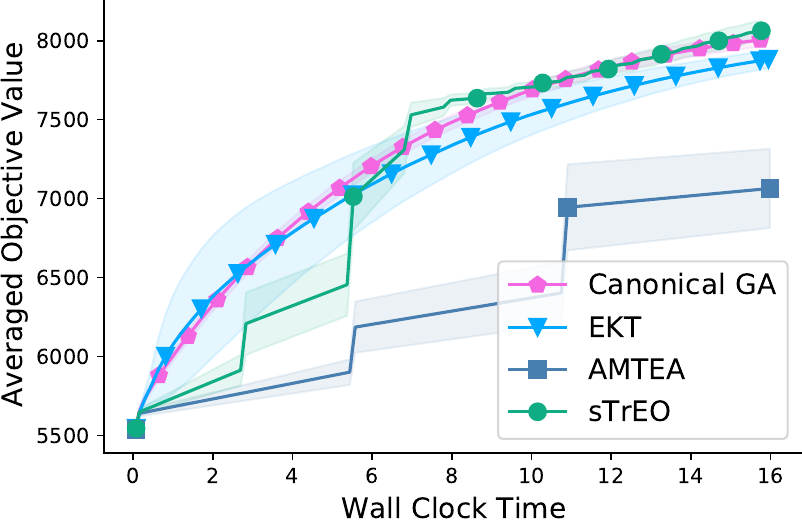}
    \end{minipage}
    }
    \subfigure[]{
    \begin{minipage}{4cm}
    \centering
    \includegraphics[width=4cm]{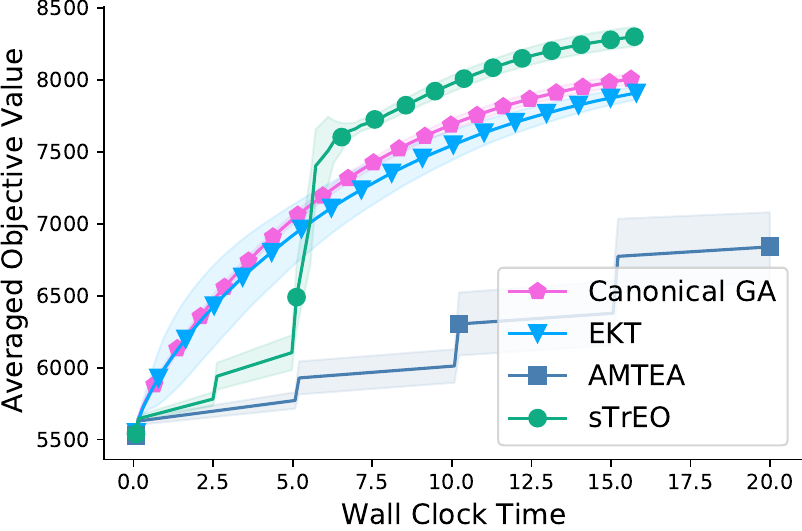}
    \end{minipage}
    }
    \caption{Convergence efficiency in terms of wall clock time in seconds with related source tasks account for (a) 0.22, (b) 0.16, (c) 0.10, and (d) 0.04 ratios of many-to-one scenario in the 0/1 Knapsack Problem.}
    \label{convergence-times-for-many-knapsack}
\end{figure*}

(2) \textbf{Time Efficiency:} 
The time efficiency of an algorithm is evaluated based on the averaged objective value achieved over the wall clock time in seconds.
Fig.~\ref{convergence-times-for-multi-knapsack} illustrates the averaged objective values of all four algorithms in solving multi-to-one knapsack problems. Accordingly, we can see that AMTEA and sTrEO obtained better performance in the early optimization process, hence demonstrating their efficiency in identifying the pertinent sources from the small number of source tasks.  
Nevertheless, in many-to-one scenarios (see Fig.~\ref{convergence-times-for-many-knapsack}), AMTEA has reported the worst performance among four algorithms as it needs to estimate the similarity between each pair of source-target tasks for each time the knowledge transfer occurs. An increase in the number of source tasks easily results in a linear scaling in the consumption of transfer costs. Moreover, sTrEO involves an incremental learning scheme to calculate task similarities online. After the early optimization stage, it obtains more competitive averaged objective values compared to CGA, EKT, and AMTEA given limited optimization time, hence verifying its online learning agility and transfer efficiency in solving knapsack problems with a ``Big Volume'' of source tasks.

\begin{figure*}[htbp]
    \centering
    \subfigure[]{
    \begin{minipage}{5cm}
    \centering
    \includegraphics[width=5cm]{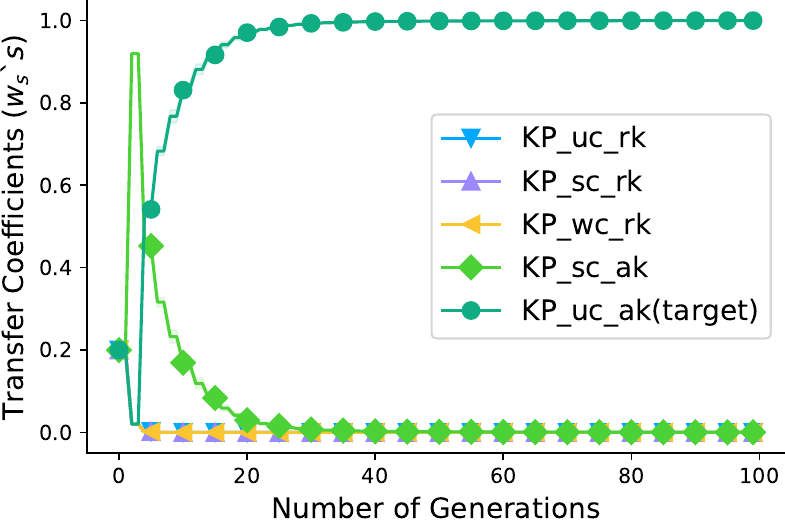}
    \end{minipage}
    }
    \subfigure[]{
    \begin{minipage}{5cm}
    \centering
    \includegraphics[width=5cm]{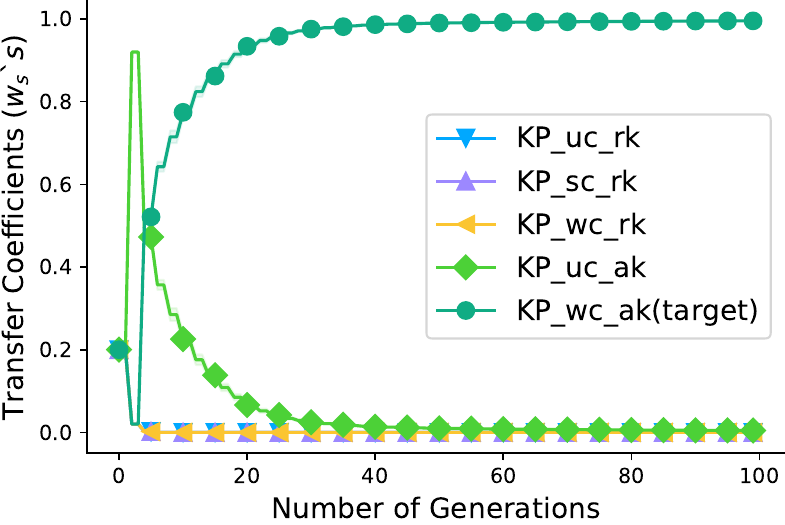}
    \end{minipage}
    }
    \subfigure[]{
    \begin{minipage}{5cm}
    \centering
    \includegraphics[width=5cm]{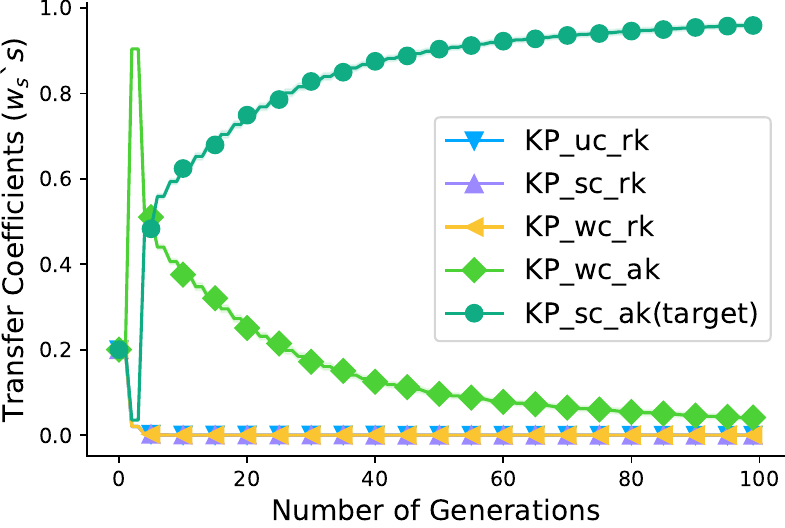}
    \end{minipage}
    }
    \caption{sTrEO's learned $w_s$'s for (a) configuration A, (b) configuration B, and (c) configuration C of multi-to-one scenario in the 0/1 Knapsack Problem.}
    \label{transfer-coefficients-for-multi-knapsack}
\end{figure*}

\begin{figure*}[htbp]
    \centering
    \subfigure[]{
    \begin{minipage}{4cm}
    \centering
    \includegraphics[width=4cm]{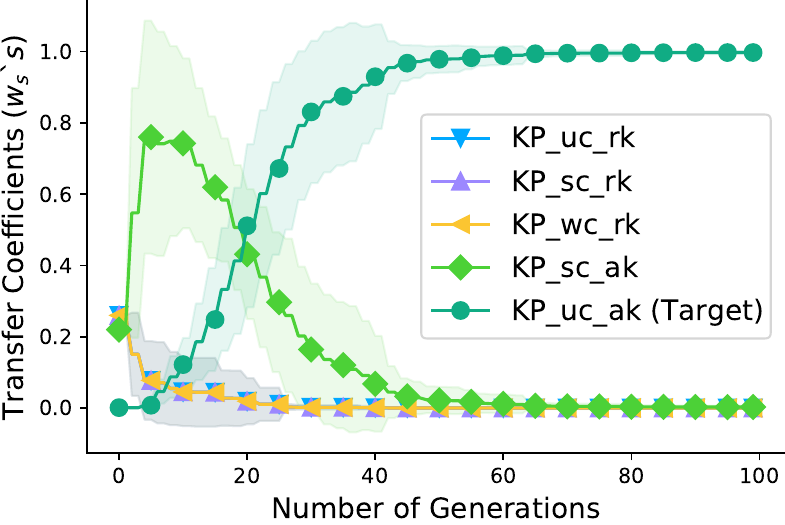}
    \end{minipage}
    }
    \subfigure[]{
    \begin{minipage}{4cm}
    \centering
    \includegraphics[width=4cm]{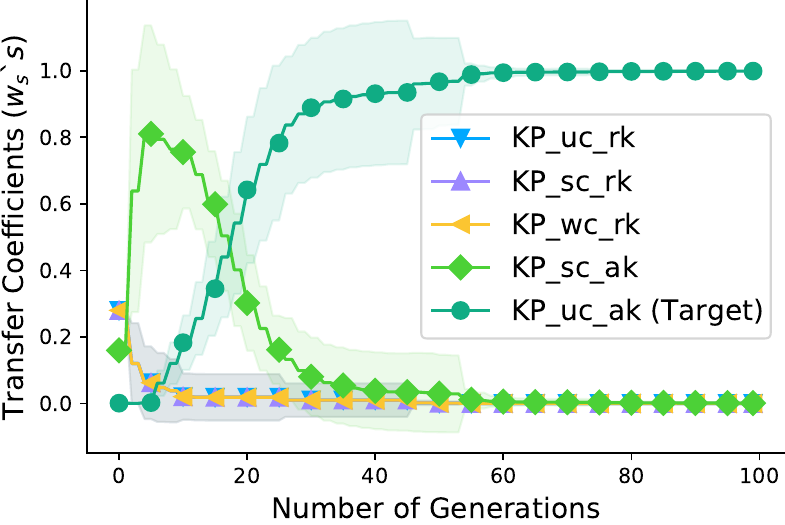}
    \end{minipage}
    }
    \subfigure[]{
    \begin{minipage}{4cm}
    \centering
    \includegraphics[width=4cm]{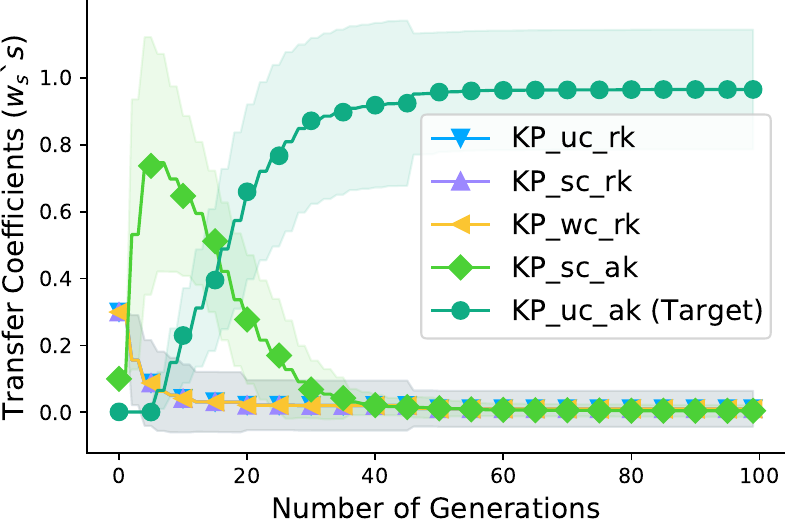}
    \end{minipage}
    }
    \subfigure[]{
    \begin{minipage}{4cm}
    \centering
    \includegraphics[width=4cm]{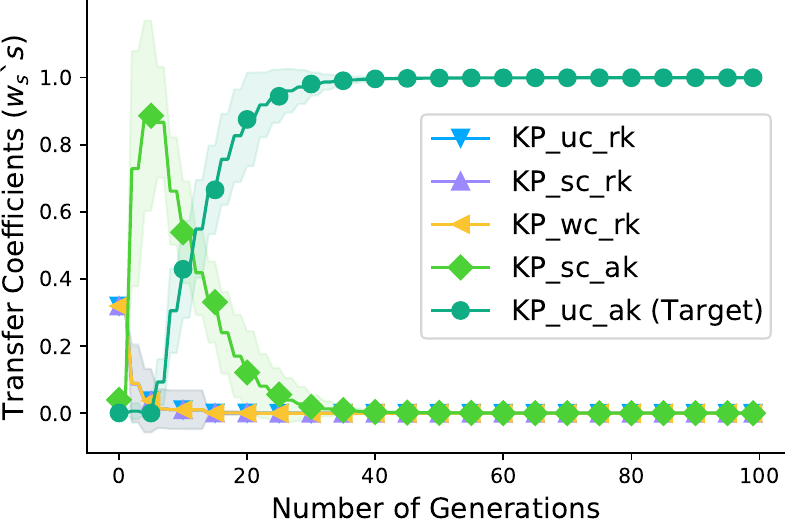}
    \end{minipage}
    }
    \caption{sTrEO's learned $w_s$'s with related source tasks account for (a) 0.22, (b) 0.16, (c) 0.10, and (d) 0.04 ratios of many-to-one scenario in the 0/1 Knapsack Problem.}
    \label{transfer-coefficients-for-many-knapsack}
\end{figure*}

(3) \textbf{Transfer Coefficients}: Particularly, the transfer coefficients of sTrEO in leveraging knowledge from different source tasks (KP\_uc\_rk, KP\_sc\_rk,  KP\_wc\_rk, KP\_sc\_ak) are depicted in Fig.~\ref{transfer-coefficients-for-multi-knapsack} and Fig.~\ref{transfer-coefficients-for-many-knapsack}. These results verify that sTrEO possesses a remarkable ability to extract valuable knowledge from the most correlated task (KP\_sc\_ak), hence speeding up the convergence of the target task of interest.

\begin{figure}
    \centering
    \subfigcapskip=5pt 
    \subfigure[Quality vs Efficiency of multi-to-one]{
    \begin{minipage}{4cm}
    \centering
    \includegraphics[width=4cm]{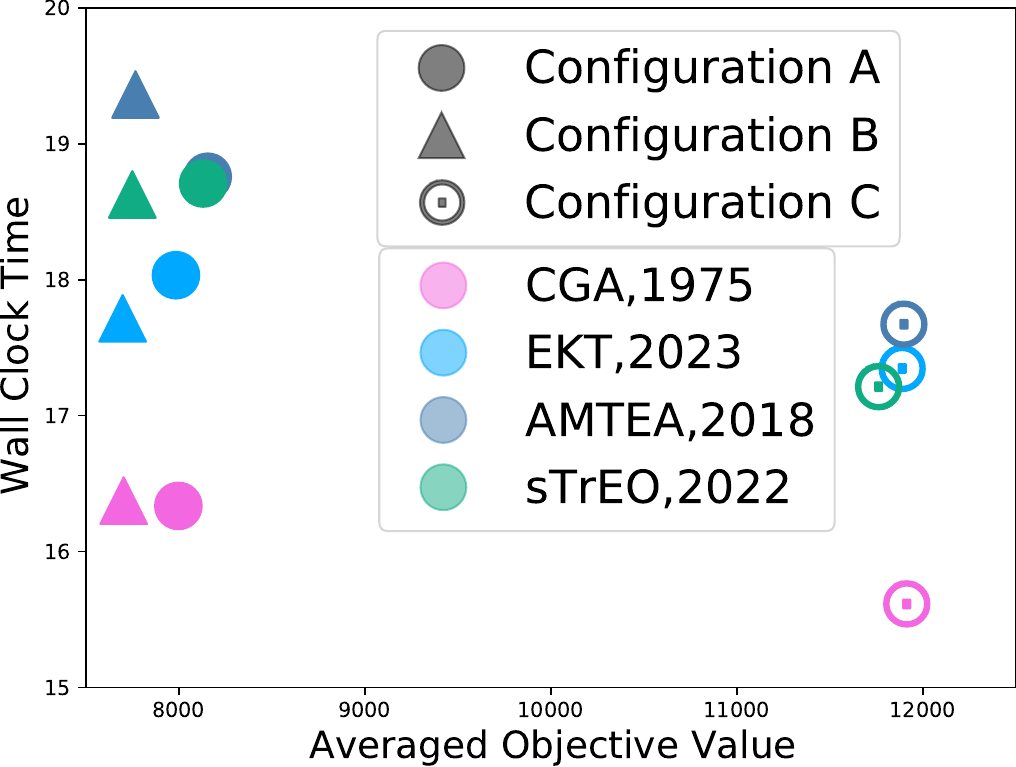}
    \end{minipage}
    }
    \subfigure[Quality vs Efficiency of many-to-one]{
    \begin{minipage}{4cm}
    \centering
    \includegraphics[width=4cm]{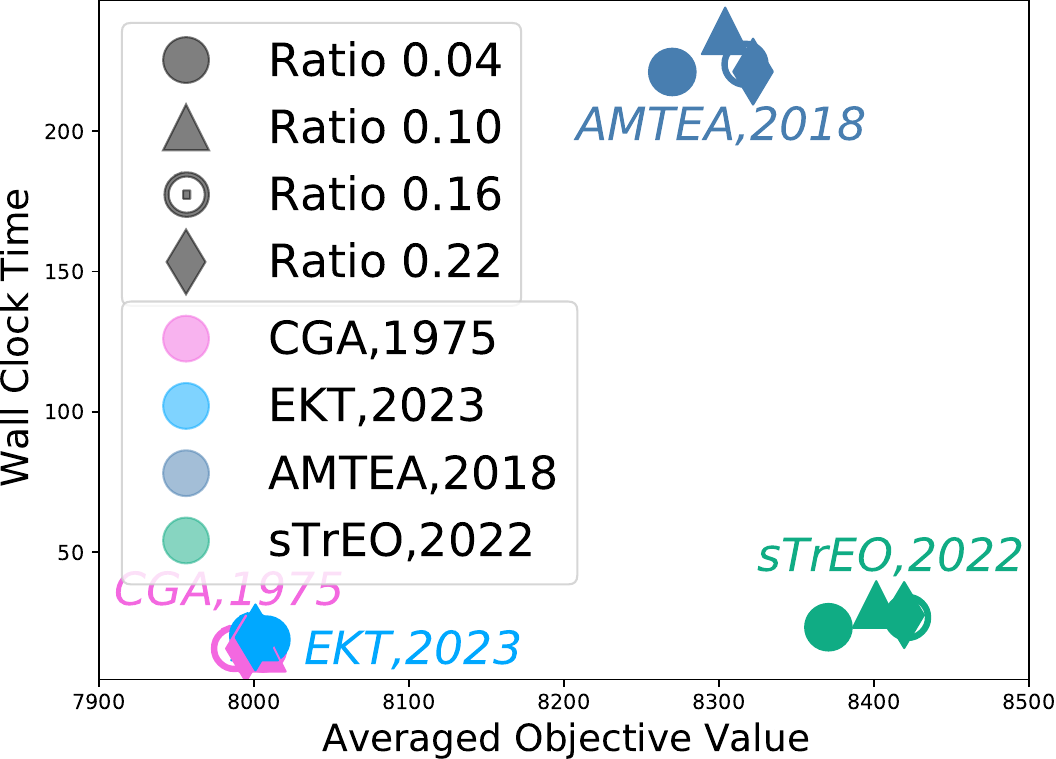}
    \end{minipage}
    }
    \caption{The Quality vs Efficiency on two objectives (averaged objective value and wall clock time) of (a) multi-to-one and (b) many-to-one scenarios in the 0/1 Knapsack Problem.}
    \label{kp-PF}
\end{figure}

(4) \textbf{Quality vs Efficiency}: Fig.~\ref{kp-PF} depicts the Quality vs Efficiency of the four algorithms in the multi-to-one and many-to-one scenarios of the knapsack problem, which represents the trade-off between the averaged objective values and time cost in terms of wall clock time after the optimization process (i.e., 5000 function evaluations). A point closer to the bottom right corner indicates better performance of the method. The results effectively validate the ``no free lunch theorem''. For instance, in configurations A and B of the multi-to-one scenario, the AMTEA algorithm achieves higher fitness than the other three algorithms but incurs a higher time overhead.

\subsection {Big Volume and Big Variety}
This section presents details of both multi-to-one and many-to-one transfer scenarios for robotic arm problems, which are characterized by ``Big Volume'' and ``Big Variety''.
After generating the heat map to analyze the correlations between
tasks (Fig.~\ref{heatmap} in Section~\ref{arm_description_correlation}), 
we designate instances with $0 < L < \sqrt{2}$ and $\alpha_{max} = 1$ (the lighter area in Fig.~\ref{heatmap}) as strongly related to the target task $\mathcal{T}_{\sqrt{2},1}$. While the instances with $0 < L < \sqrt{2}$ and $0.18 < \alpha_{max} < 0.26$ (the darker area in Fig.~\ref{heatmap}) are considered to be weakly related to the target task. 
Furthermore, to test the performance of transfer optimization in handling source tasks with ``Big Variety'' characteristics, we re-configure the dimensionality of the source tasks (i.e., the number of joints) as $d_s=\psi * d_T$, and the robotic arm length as $L_s=L_T / \psi$, where $\psi$ is a scaling factor randomly generated from $\{2,3,...,9\}$. $d_T$ and $L_T$ represent the dimensionality and the robotic arm length of the target task, respectively. In this setting, the relevance between source and target tasks can still be retained.

\emph{Multi-to-one Problem:} In this scenario, we consider two configurations with 10 joints and 20 joints, respectively. For each configuration, we select three weakly related instances and one strongly related instance as source tasks, together with the target task to form the multi-to-one problem.

\emph{Many-to-one Problem:} 
In this scenario, both configurations with 10 joints and 20 joints are taken into account. we have set up 1000 source tasks, with 20 of them being strongly related to the target task, constituting a proportion of 0.02. 

\emph{Experimental Settings:} In the planar robotic arm problem, we assess and compare the performance of multiple algorithms: (\textit{i}) Canonical Genetic Algorithm~\cite{cga} (CGA), (\textit{ii}) Evolutionary Knowledge Transfer~\cite{EKT} (EKT),  (\textit{iii}) MultiSource Selective Transfer Optimization~\cite{MSSTO} (MSSTO, a model-free transfer method with a linear mapping), (\textit{iv}) Direction Vector-based Transfer Evolutionary Optimization~\cite{MFEA-DV} (DVTrEO, a direction vector-based transfer algorithm modified from MFEA-DV), (\textit{v}) Affine Transformation-enhanced Transfer Evolutionary Optimization~\cite{AT-MFEA} (ATTrEO, a mapping-based transferability enhancement algorithm modified from AT-MFEA), (\textit{vi}) Adaptive Model-based Transfer Evolutionary Algorithm~\cite{curbing-AMTEA} (AMTEA), and (\textit{vii}) scalable Transfer Evolutionary Optimization~\cite{sTrEO} (sTrEO).
Note that certain algorithms in TrEO, are not specifically designed to handle heterogeneity in the search space. For such algorithms, a simple mapping method is applied. If the dimensionality of the source task is greater than that of the target task, we clip the optimized solutions in the source task. Otherwise, we fill the optimized solutions with random values. 

All source tasks are optimized by a continuous CGA~\cite{cga} at first. Then, the solutions of the first and last population are archived to offer knowledge. Baseline results of the above algorithms are provided. Moreover, the following settings are adopted in the experiments:

\begin{enumerate}
    \item Representation: Real-value coded in the range [0, 1].
    \item Repetition: 30.
    \item Population size: 50.
    \item Maximum function evaluations: 5000.
    \item Evolutionary operators:
    \begin{enumerate}
        \item Simulated binary crossover~\cite{SBX, crossover-operators} with probability $p_c$=1 and distribution index $\eta_c$ = 10.
        \item Polynomial-based mutation~\cite{polynomial-mutation} with probability $p_m=1/d$ ($d$ is the dimensionality of the target optimization problem) and distribution index $\eta_m$=10.
    \end{enumerate}
    \item For model-based transfer algorithm(s):
    \begin{enumerate}
    \item Probabilistic model: Multivariate Gaussian distribution~\cite{mvarnorm}.
        \item Transfer interval: 2.
    \end{enumerate}
\end{enumerate} 

\emph{Performance Metric:} Similar to knapsack problems, we take the averaged objective value, performance score, and wall clock time into account in the planar arm problem. Readers can refer to \ref{knapsack-metrics} for more details. 

\emph{Comparison of Results and Analysis:}

\begin{figure}
    \centering
    \subfigure[]{
    \begin{minipage}{4cm}
    \centering
    \includegraphics[width=4cm]{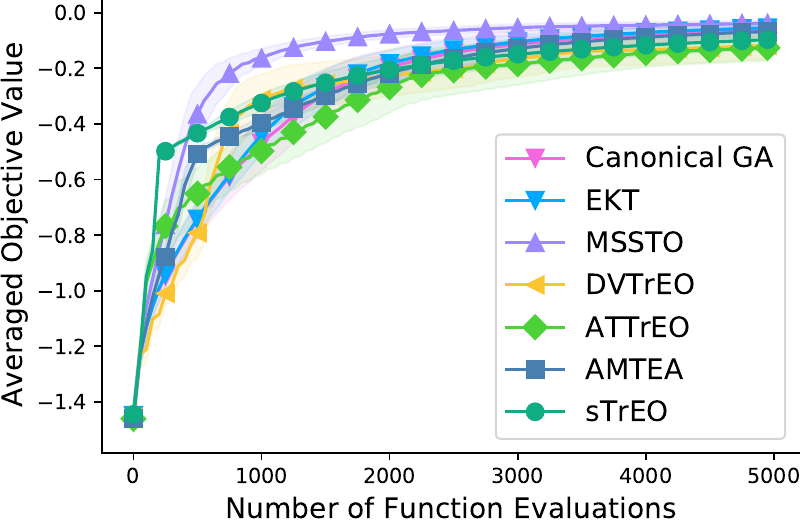}
    \end{minipage}
    }
    \subfigure[]{
    \begin{minipage}{4cm}
    \centering
    \includegraphics[width=4cm]{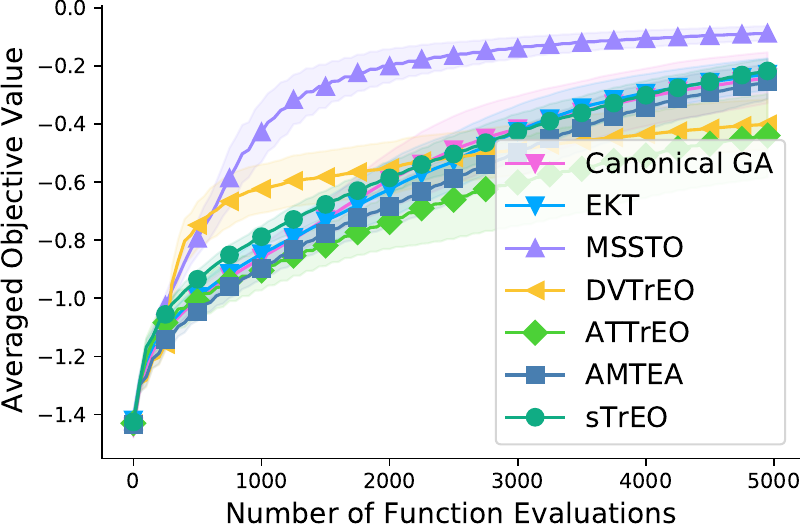}
    \end{minipage}
    }
    \caption{Convergence trends for (a) 10 joints and (b) 20 joints of multi-to-one scenario in the Planar Robotic Arm Problem.}
    \label{convergence-trends-for-multi-arm}
\end{figure}

\begin{figure}
    \centering
    \subfigure[]{
    \begin{minipage}{4cm}
    \centering
    \includegraphics[width=4cm]{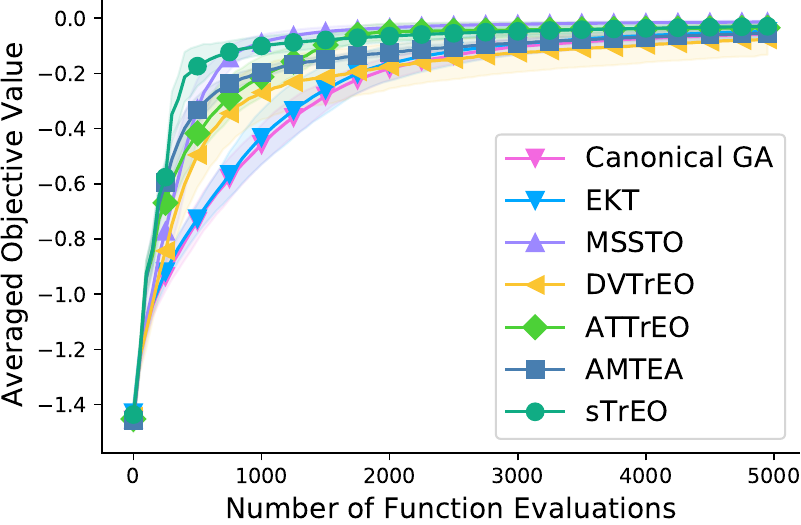}
    \end{minipage}
    }
    \subfigure[]{
    \begin{minipage}{4cm}
    \centering
    \includegraphics[width=4cm]{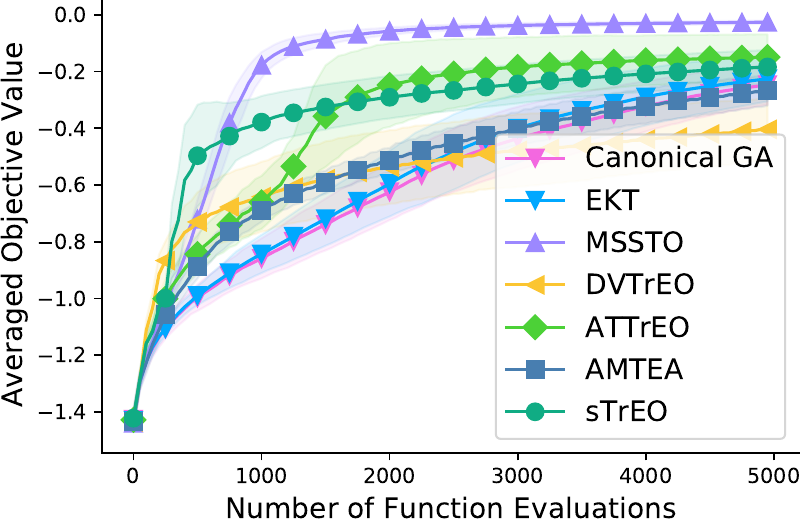}
    \end{minipage}
    }
    \caption{Convergence trends for (a) 10 joints and (b) 20 joints of many-to-one scenario in the Planar Robotic Arm Problem.}
    \label{convergence-trends-for-many-arm}
\end{figure}

(1) \textbf{Data Efficiency}: Tables~\ref{multi-arm-results-objective-value},~\ref{multi-arm-results-performance-score},~\ref{many-arm-results-objective-value} and~\ref{many-arm-results-performance-score} summarize the averaged objective values and performance scores of CGA, EKT, MSSTO, DVTrEO, ATTrEO, AMTEA, and sTrEO after 5000 function evaluations. The best results are shown in {\bf boldface}.  Their corresponding convergence trends for the averaged objective values are depicted in Fig.~\ref{convergence-trends-for-multi-arm} and Fig.~\ref{convergence-trends-for-many-arm}, respectively. 
Due to the ``Big Variety'' nature of the planar robotic arm problem (reflected in the dimensional mismatch between the source tasks and the target task), TrEO algorithms, which extract knowledge from the source tasks and directly utilize it without any mapping, perform poorly in this problem. It can be seen that the MSSTO algorithm, which essentially builds a mapping/adaptation mechanism based on a single-layer denoising autoencoder~\cite{autoencoding}, outperforms all the other transfer optimization algorithms in terms of convergence speed and averaged objective values after the optimization process, thus confirming its superiority in solving the planar arm problem with a large volume and a large variety of source task instances.

\begin{table*}[!htbp] 
    \renewcommand\arraystretch{1.5}
    \centering
    \caption{Averaged objective values of CGA, EKT, MSSTO, DVTrEO, ATTrEO, AMTEA, and sTrEO in multi-to-one scenarios of the Planar Robotic Arm problem with 10 and 20 joints, respectively. The best are shown in \textbf{boldface}.}
    \setlength{\tabcolsep}{6mm}{
        \scalebox{0.9}{
        \begin{tabular}{cccccccc}
            \toprule 
            \multicolumn{1}{c}{Joint Number} & CGA & EKT & MSSTO & DVTrEO & ATTrEO & AMTEA & sTrEO \\  
            \hline 
            \multicolumn{1}{c}{10} & -0.0592 & -0.0541 &  \textbf{-0.0384} & -0.1181 & -0.1267 & -0.0678 & -0.0972 \\  
            \multicolumn{1}{c}{20} & -0.2409 & -0.2284 & \textbf{-0.0872} & -0.4008 & -0.4394 & -0.2532 & -0.2171 \\
            \bottomrule
        \end{tabular}
        }
        }
    \label{multi-arm-results-objective-value}
\end{table*}

\begin{table*}[!htbp] 
    \renewcommand\arraystretch{1.5}
    \centering
    \caption{Performance scores of CGA, EKT, MSSTO, DVTrEO, ATTrEO, AMTEA, and sTrEO in multi-to-one scenarios of Planar Robotic Arm problem with 10 and 20 joints, respectively. The best are shown in \textbf{boldface}.}
    \setlength{\tabcolsep}{6mm}{
        \scalebox{0.9}{
        \begin{tabular}{cccccccc}
            \toprule 
            \multicolumn{1}{c}{Joint Number} & CGA & EKT & MSSTO & DVTrEO & ATTrEO & AMTEA & sTrEO \\  
            \hline 
            \multicolumn{1}{c}{10} &  0.4398 & 0.5574 & \textbf{0.9218} & -0.9228 & -0.7978 & 0.2408 & -0.4392 \\  
            \multicolumn{1}{c}{20} &  0.1478 & 0.2453 & \textbf{1.3397} & -1.0915 & -1.0268 & 0.0525 & 0.3329\\
            \bottomrule 
        \end{tabular}
        }
        }
    \label{multi-arm-results-performance-score}
\end{table*}

\begin{table*}[!htbp] 
    \renewcommand\arraystretch{1.5}
    \centering
    \caption{Averaged objective values of CGA, EKT, MSSTO, DVTrEO, ATTrEO, AMTEA, and sTrEO in many-to-one scenarios of the Planar Robotic Arm Problem with 10 and 20 joints, respectively. The best are shown in \textbf{boldface}.}
    \setlength{\tabcolsep}{6mm}{
        \scalebox{0.9}{
        \begin{tabular}{cccccccc}
            \toprule 
            \multicolumn{1}{c}{Joint Number} & CGA & EKT & MSSTO & DVTrEO & ATTrEO & AMTEA & sTrEO \\ 
            \hline 
            \multicolumn{1}{c}{10} &  -0.0570 & -0.0481 & \textbf{-0.0138} & -0.0766 & -0.0347 & -0.0542 & -0.0294 \\  
            \multicolumn{1}{c}{20} &  -0.2463 & -0.2254 & \textbf{-0.0256} & -0.4036 & -0.1488 & -0.2662 & -0.1835\\
            \bottomrule 
        \end{tabular}
        }
        }
    \label{many-arm-results-objective-value}
\end{table*} 

\begin{table*}[!htbp] 
    \renewcommand\arraystretch{1.5}
    \centering
    \caption{Performance scores of CGA, EKT, MSSTO, DVTrEO, ATTrEO, AMTEA, and sTrEO in many-to-one scenarios of the Planar Robotic Arm Problem with 10 and 20 joints, respectively. The best are shown in \textbf{boldface}.}
    \setlength{\tabcolsep}{6mm}{
        \scalebox{0.9}{
        \begin{tabular}{cccccccc}
            \toprule 
            \multicolumn{1}{c}{Joint Number} & CGA & EKT & MSSTO & DVTrEO & ATTrEO & AMTEA & sTrEO \\  
            \hline 
            \multicolumn{1}{c}{10} &  -0.3917 & -0.1178 & \textbf{0.9373} & -0.9958 & 0.4186 & -0.307 & 0.4566 \\ 
            \multicolumn{1}{c}{20} &  -0.2670 & -0.1050 & \textbf{1.4425} & -1.4853 & 0.6163 & -0.4210 & 0.2196 \\
            \bottomrule 
        \end{tabular}
        }
        }
    \label{many-arm-results-performance-score}
\end{table*} 

% arm many generation

\begin{figure}
    \centering
    \subfigure[]{
    \begin{minipage}{4cm}
    \centering
    \includegraphics[width=4cm]{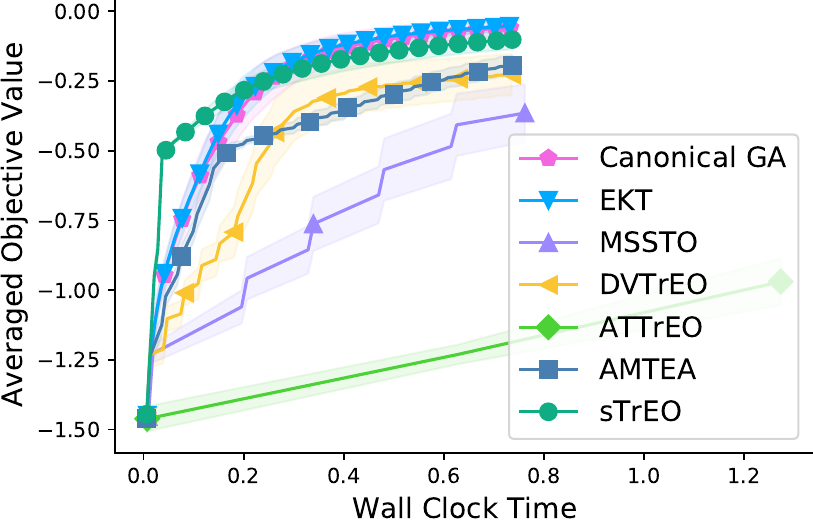}
    \end{minipage}
    }
    \subfigure[]{
    \begin{minipage}{4cm}
    \centering
    \includegraphics[width=4cm]{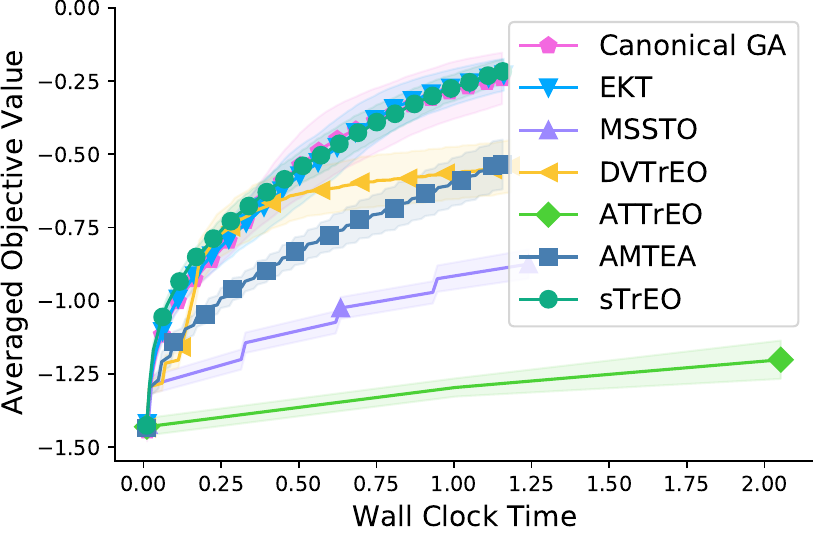}
    \end{minipage}
    }
    \caption{Convergence efficiency in terms of wall clock time in seconds in multi-to-one scenarios of the Planar Robotic Arm Problem with (a) 10 joints and (b) 20 joints, respectively.}
    \label{convergence-times-for-multi-arm}
\end{figure}

\begin{figure}
    \centering
    \subfigure[]{
    \begin{minipage}{4cm}
    \centering
    \includegraphics[width=4cm]{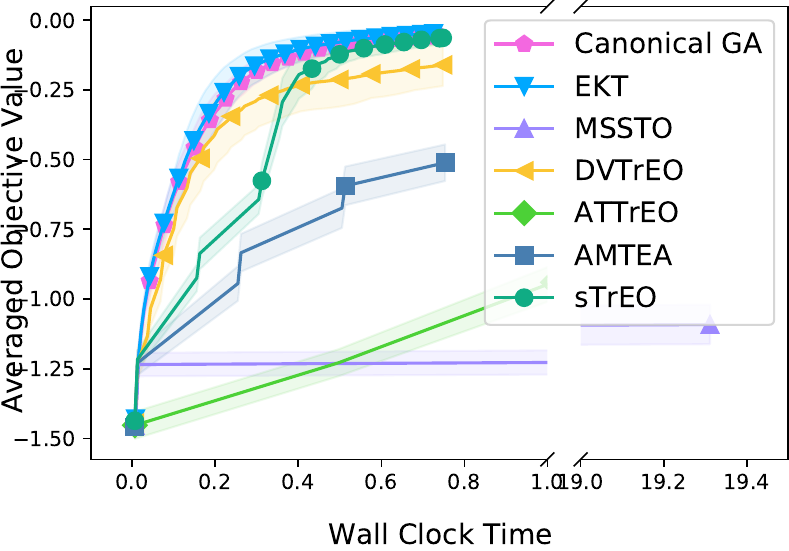}
    \end{minipage}
    }
    \subfigure[]{
    \begin{minipage}{4cm}
    \centering
    \includegraphics[width=4cm]{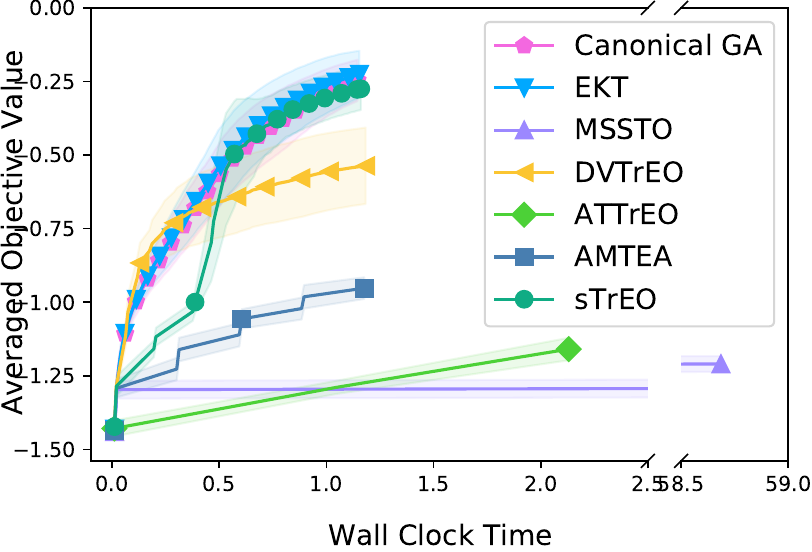}
    \end{minipage}
    }
    \caption{Convergence efficiency in terms of wall clock time in seconds in many-to-one scenarios of the Planar Robotic Arm Problem with (a) 10 joints and (b) 20 joints, respectively.}
    \label{convergence-times-for-many-arm}
\end{figure}

(2) \textbf{Time Efficiency}: The time efficiency in the planar arm problem is measured in the same way as in the knapsack problem. As can be seen in Figs.~\ref{convergence-times-for-multi-arm} and~\ref{convergence-times-for-many-arm}, AMTEA, ATTrEO, and MSSTO are significantly more time-consuming because they all adopted a polling-based source-target similarity measurement for each time of the knowledge transfer, especially in many-to-one scenarios with a large number of source tasks. Furthermore, compared to AMTEA and ATTrEO, MSSTO has a more expensive time complexity as it uses all generated solutions to learn the auto-encoder based source-target mapping.

\begin{figure}
    \centering
    \subfigcapskip=5pt 
    \subfigure[Quality vs Efficiency of multi-to-one]{
    \begin{minipage}{4cm}
    \centering
    \includegraphics[width=4cm]{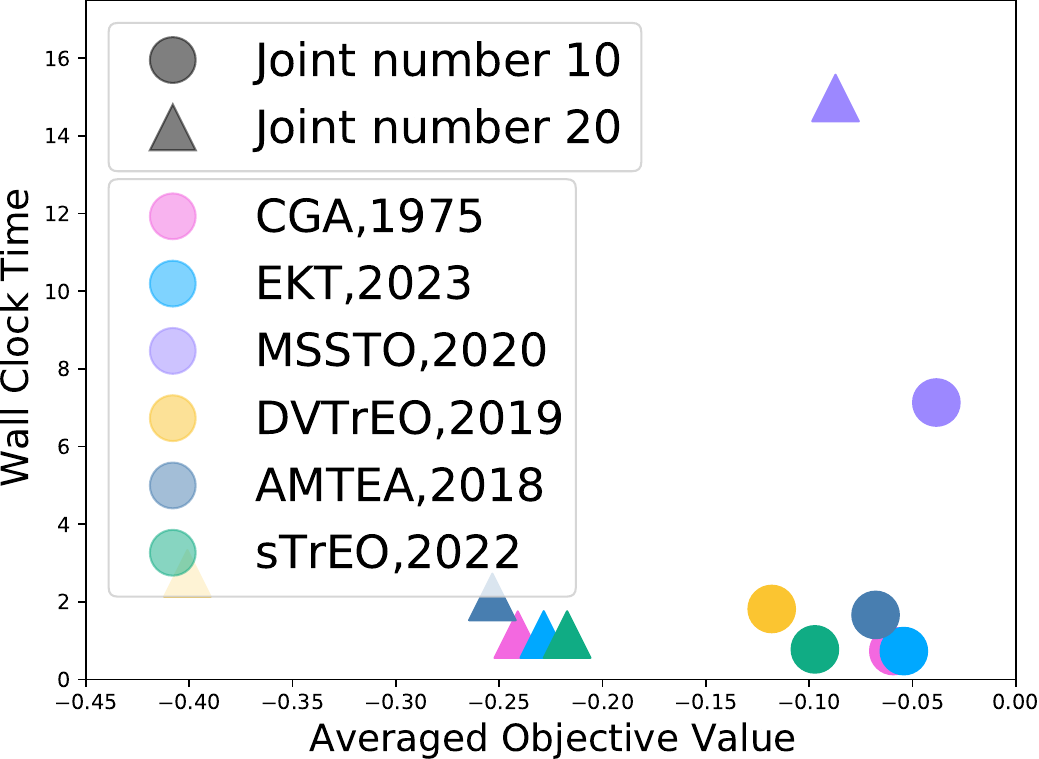}
    \end{minipage}
    }
    \subfigure[Quality vs Efficiency of many-to-one]{
    \begin{minipage}{4cm}
    \centering
    \includegraphics[width=4cm]{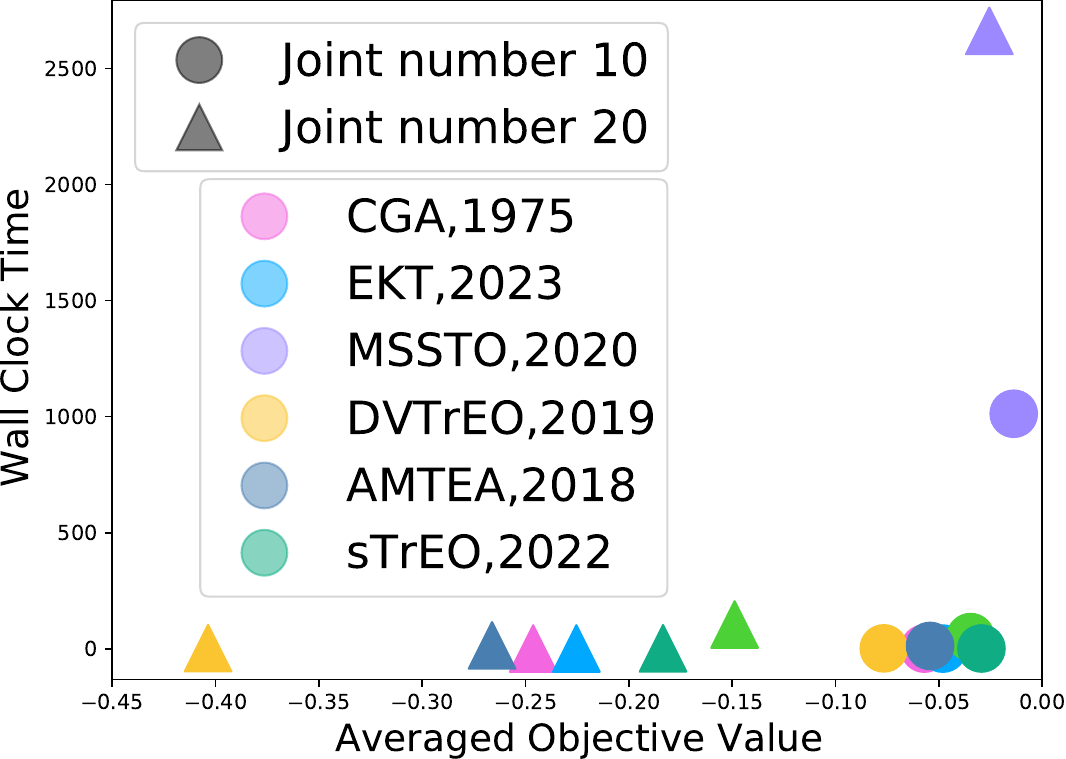}
    \end{minipage}
    }
    \caption{The Quality vs Efficiency on two objectives (averaged objective value and wall clock time) in (a) multi-to-one and (b) many-to-one scenarios of the Planar Robotic Arm Problem after the optimization process.}
    \label{arm-PF}
\end{figure}

(3) \textbf{Quality vs Efficiency}: Fig.~\ref{arm-PF} shows the Quality vs Efficiency of CGA, EKT, MSSTO, DVTrEO, AMTEA, and sTrEO in multi-to-one and many-to-one scenarios of the planar arm problem after 5000 function evaluations. A point closer to the lower right corner is preferred. Note that ATTrEO requires significantly more wall clock time to complete the optimization process than other algorithms and is therefore excluded from the figures. We can see that MSSTO performs best in terms of average objective value, but takes a much longer time to complete 5000 function evaluations. Thus, the ``no free lunch theorem'' is well proven here.

\subsection {Big Volume and Big Velocity}
Minimalistic Attacks exhibit the ``Big Volume'' and ``Big Velocity'' characteristics and their detailed experimental configuration and analysis are provided as follows.
Specifically, we consider the ACKTR~\cite{acktr} as our essential reinforcement learning policy and employ three games: BeamRider, Qbert, and Seaquest as the baseline environments. For each environment, the target task is first defined. Then, we pick frames with strong and weak correlations to the target task as source tasks to construct our minimalistic attack problems. More details are discussed as follows.

This study focuses solely on the many-to-one transfer scenario since the agent in Atari games can interact with the environment and generate hundreds, or even thousands, of frames in a relatively short amount of time. Specifically, we set the number of source tasks to 1000. For each game, we consider three configurations where the related source tasks account for 0.24, 0.14, and 0.04 of all source tasks.
This setting hence creates a ``Big Volume'' scenario where the knowledge from 1000 source tasks can be transferred to a single target task (i.e., 1000-to-1), with varying ratios of related tasks. Meanwhile, in minimalistic attacks, the transfer optimization algorithm has to launch a successful attack within a tight time window (i.e., before the keyframe vanishes), hence posing a significant challenge in handling the ``Big Velocity'' of source task instances.

\emph{Experimental Settings:}
In minimalistic attacks, we consider a setup with 4 attacked pixels. Each pixel is encoded using three consecutive gene positions, corresponding to its x-coordinate, y-coordinate, and the perturbation added to those coordinates. This formulation results in a problem dimensionality of 12. Likewise, we employ (\textit{i}) Canonical Genetic Algorithm~\cite{cga} (CGA), (\textit{ii}) Evolutionary Knowledge Transfer~\cite{EKT} (EKT),  (\textit{iii}) MultiSource Selective Transfer Optimization~\cite{MSSTO} (MSSTO), (\textit{iv}) Direction Vector-based Transfer Evolutionary Optimization~\cite{MFEA-DV} (DVTrEO), (\textit{v}) Affine Transformation-enhanced Transfer Evolutionary Optimization~\cite{AT-MFEA} (ATTrEO), (\textit{vi}) Adaptive Model-based Transfer Evolutionary Algorithm~\cite{curbing-AMTEA} (AMTEA), and (\textit{vii}) scalable Transfer Evolutionary Optimization~\cite{sTrEO} (sTrEO) on this problem. 

All source tasks are optimized by CGA at first and the solutions of the first and last population are archived. The experimental parameters and configurations are as follows:

\begin{enumerate}
    \item Representation: Real-valued coded in the range [0, 1].
    \item Repetition: 30.
    \item Population size: 10.
    \item Maximum function evaluations: 1000.
    \item Evolutionary operators:
    \begin{enumerate}
        \item Single-point crossover~\cite{crossover-operators} with probability $p_c=1$.
        \item Gaussian mutation with probability $p_m=0.1$, mean $\mu=0$, and standard deviation $\sigma=0.5$.
    \end{enumerate}
    \item For model-based transfer algorithm(s):
    \begin{enumerate}
    \item Probabilistic model: Multivariate Gaussian distribution~\cite{mvarnorm}.
        \item Transfer interval: 2.
    \end{enumerate}
\end{enumerate}

\emph{Performance Metric:} For minimalistic attacks, we employ the generation number at which the loss value first exceeds 0 (labeled as $Gen_{gt0}$) to evaluate the performance of CGA, EKT, MSSTO, DVTrEO, ATTrEO, AMTEA, and sTrEO. This criterion is employed because a positive loss value indicates a successful attack. Therefore, a smaller $Gen_{gt0}$ indicates that the algorithm can launch a successful attack more quickly.\\

\emph{Comparison Results and Analysis:} 

\begin{figure*}[htbp]
    \centering
    \subfigure[BeamRide environment with ratio of 0.24]{
    \begin{minipage}{5cm}
    \centering
    \includegraphics[width=5cm]{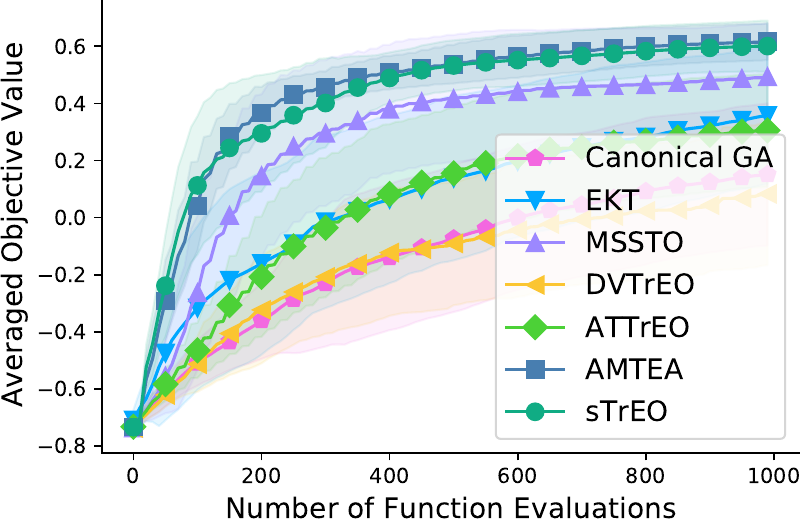}
    \end{minipage}
    }
    \subfigure[BeamRide environment with ratio of 0.14]{
    \begin{minipage}{5cm}
    \centering
    \includegraphics[width=5cm]{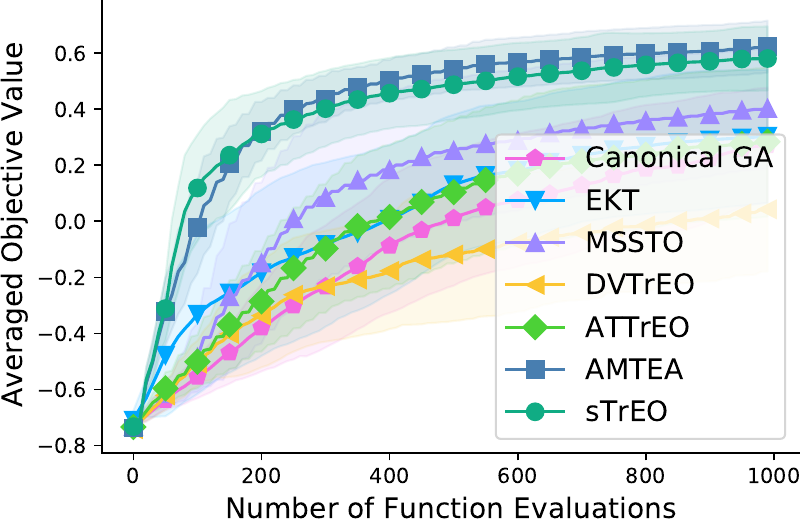}
    \end{minipage}
    }
    \subfigure[BeamRide environment with ratio of 0.04]{
    \begin{minipage}{5cm}
    \centering
    \includegraphics[width=5cm]{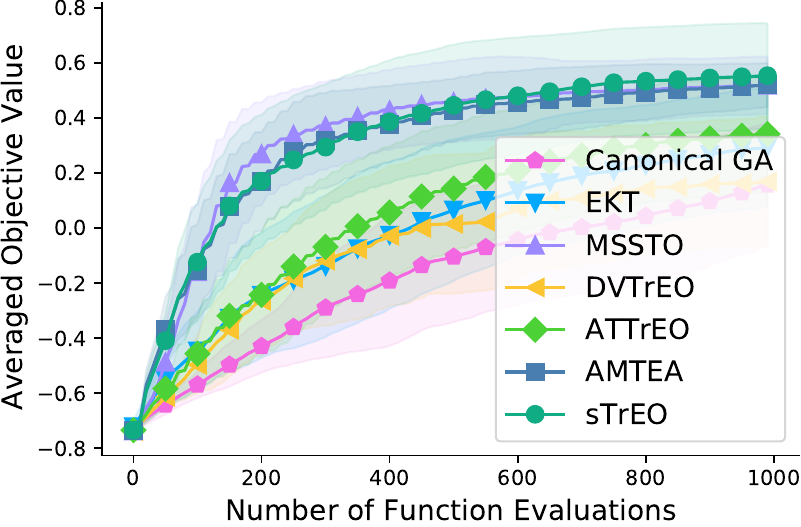}
    \end{minipage}
    }

    \centering
    \subfigure[Qbert environment with ratio of 0.24]{
    \begin{minipage}{5cm}
    \centering
    \includegraphics[width=5cm]{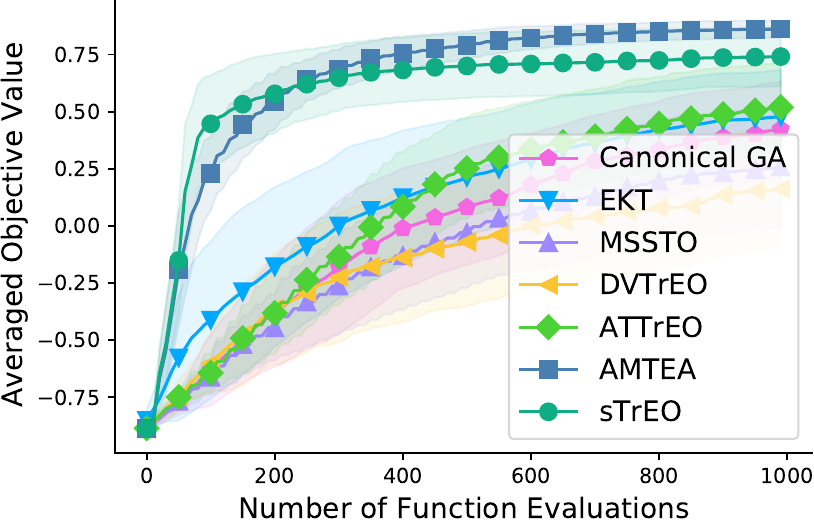}
    \end{minipage}
    }
    \subfigure[Qbert environment with ratio of 0.14]{
    \begin{minipage}{5cm}
    \centering
    \includegraphics[width=5cm]{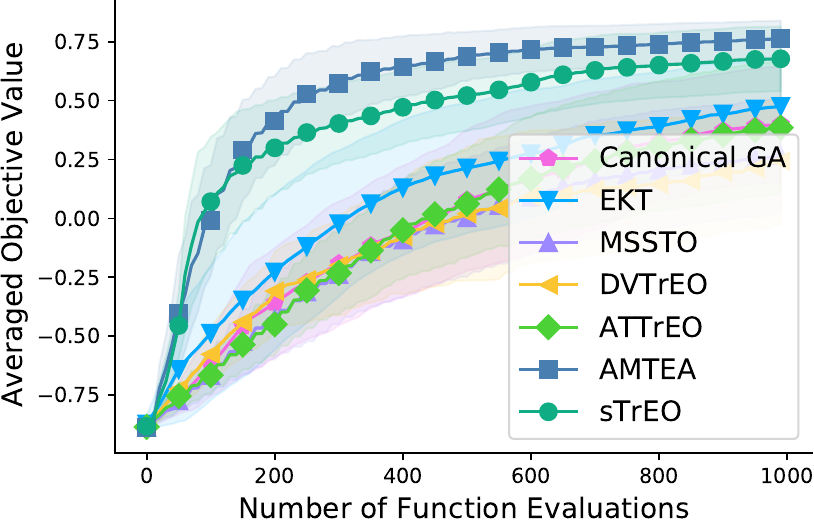}
    \end{minipage}
    }
    \subfigure[Qbert environment with ratio of 0.04]{
    \begin{minipage}{5cm}
    \centering
    \includegraphics[width=5cm]{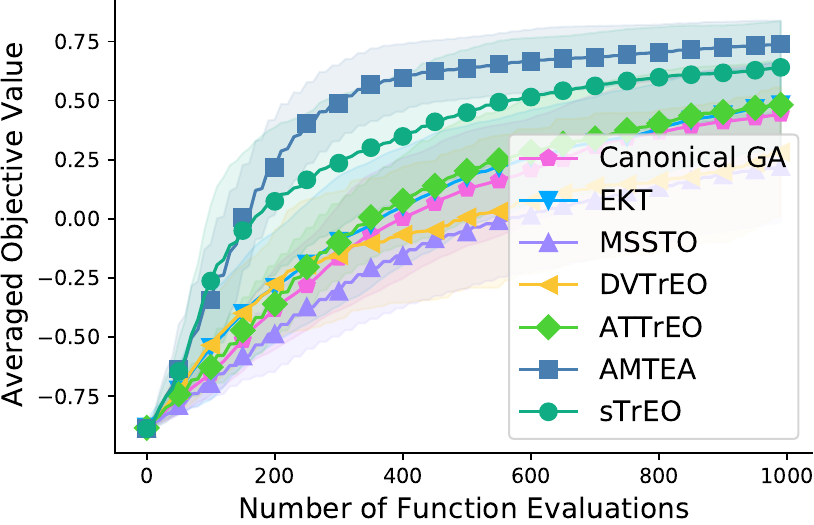}
    \end{minipage}
    }

    \centering
    \subfigure[Seaquest environment with ratio of 0.24]{
    \begin{minipage}{5cm}
    \centering
    \includegraphics[width=5cm]{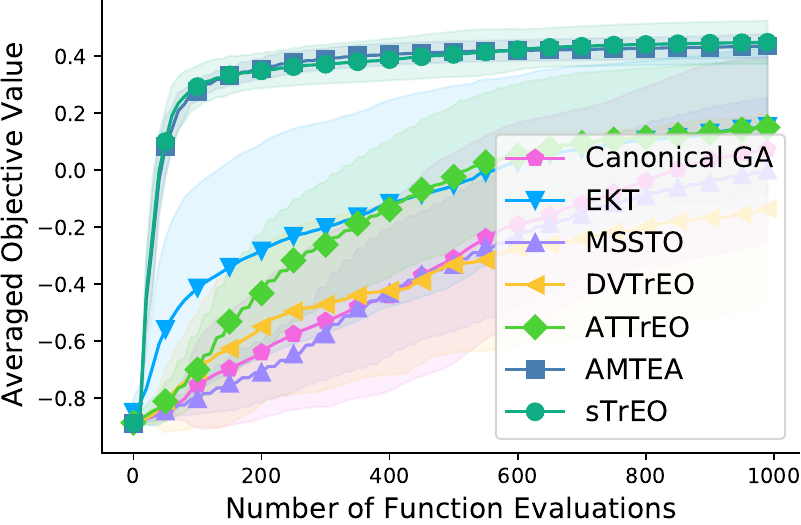}
    \end{minipage}
    }
    \subfigure[Seaquest environment with ratio of 0.14]{
    \begin{minipage}{5cm}
    \centering
    \includegraphics[width=5cm]{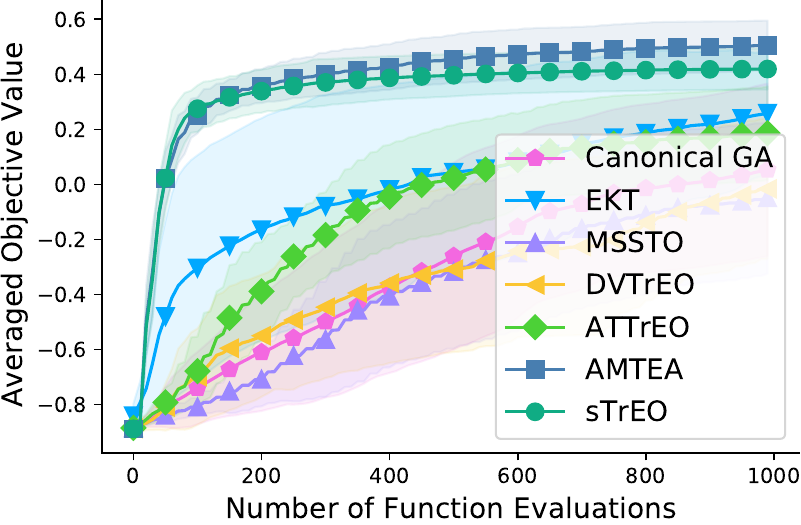}
    \end{minipage}
    }
    \subfigure[Seaquest environment with ratio of 0.04]{
    \begin{minipage}{5cm}
    \centering
    \includegraphics[width=5cm]{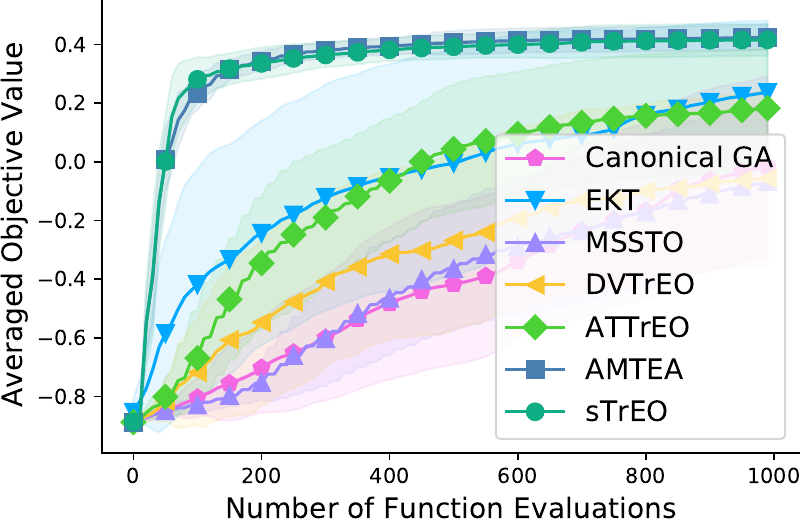}
    \end{minipage}
    }
    \caption{Convergence trends for three different game environments with three different ratios (in descending order of the proportion of related source tasks) in Minimalistic Attacks.}
    \label{convergence-trends-for-many-attack}
\end{figure*}

(1) \textbf{Data Efficiency}: Table~\ref{many-attack-results} shows the $Gen_{gt0}$ values of CGA, EKT, MSSTO, DVTrEO, ATTrEO, AMTEA, and sTrEO on attacking three games in minimalistic attacks. 
The outcomes highlighted in {\bf boldface} are the most favorable. The results show that the performance of various TrEO algorithms is distinct from each other. Among all methods, sTrEO and AMTEA achieve better $Gen_{gt0}$ values than the other algorithms, indicating their superiority in generating successful attacks with fewer evaluations. Fig.~\ref{convergence-trends-for-many-attack} shows the convergence trends of the averaged objective values obtained by seven algorithms in three games. Remarkably, sTrEO and AMTEA consistently perform well in all scenarios due to their effectiveness and efficiency in extracting useful knowledge from source tasks, enabling rapid growth and quick convergence. In particular, as the ratio of correlated source tasks decreases (i.e. from 0.24 to 0.04), the $Gen_{gt0}$ values reported by algorithms generally increase. This highlights the increasing difficulty of knowledge transfer as the ratio of related source tasks decreases.

\begin{table*}[!htbp] 
    \centering
    \caption{The $Gen_{gt0}$ of CGA, EKT, MSSTO, DVTrEO, ATTrEO, AMTEA, and sTrEO for three different game environments with three different ratios (in descending order of the proportion of related source tasks) in Minimalistic Attacks. The best results are shown in \textbf{boldface}.}
    \scalebox{0.95}{
        \begin{tabular}{cccccccccc} 
            \toprule 
            \multicolumn{2}{c}{Game Environment} &
            \multicolumn{1}{c}{Ratio} &
            CGA & EKT & MSSTO & DVTrEO & ATTrEO & AMTEA & sTrEO\\  
            \hline 
            
            \multicolumn{2}{c}{\multirow{3}*{BeamRider}} & \multicolumn{1}{c}{0.24} & 61 & 33 & 15 & 74 & 33 & 9 & \textbf{8} \\
    		\multicolumn{2}{c}{~} & \multicolumn{1}{c}{0.14} & 49 & 40 & 25 & 84 & 38 & 11 & \textbf{8} \\
    		\multicolumn{2}{c}{~} & \multicolumn{1}{c}{0.04} & 69 & 44 & 18 & 45 & 35 & 15 & \textbf{13} \\
            \hline 
            
            \multicolumn{2}{c}{\multirow{3}*{Qbert}} & \multicolumn{1}{c}{0.24} & 42 & 30 & 53 & 60 & 37 & 7 & \textbf{6} \\
    		\multicolumn{2}{c}{~} & \multicolumn{1}{c}{0.14} & 45 & 32 & 49 & 48 & 45 & 11 & \textbf{9} \\
    		\multicolumn{2}{c}{~} & \multicolumn{1}{c}{0.04} & 40 & 37 & 57 & 50 & 35 & \textbf{15} & 17 \\
            \hline 
            
            \multicolumn{2}{c}{\multirow{3}*{Seaquest}} & \multicolumn{1}{c}{0.24} & 85 & 57 & - & - & 53 & \textbf{5} & \textbf{5} \\
    		\multicolumn{2}{c}{~} & \multicolumn{1}{c}{0.14} & 86 & 43 & - & - & 47 & \textbf{5} & \textbf{5} \\
    		\multicolumn{2}{c}{~} & \multicolumn{1}{c}{0.04} & - & 51 & - & - & 45 & \textbf{5} & \textbf{5} \\
    
            \bottomrule 
        \end{tabular}
    }
    \label{many-attack-results}
\end{table*}

\begin{figure}
    \centering
    \includegraphics[width=0.4\textwidth]{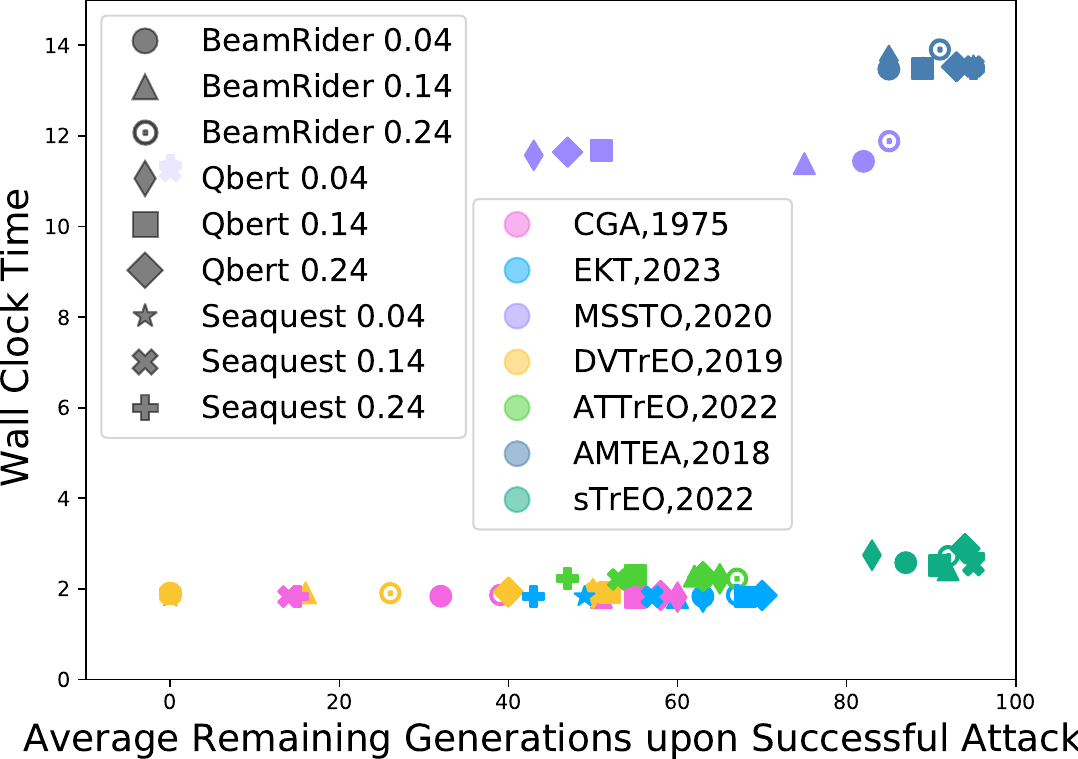}
    \caption{The Quality vs Efficiency on two objectives (averaged objective value and wall clock time) for each different game environment in Minimalistic Attacks. sTrEO and EKT are on the front and they dominate the rest of the algorithms}
    \label{attack-many-PF}
\end{figure}

(2) \textbf{Quality vs Efficiency}: Fig.~\ref{attack-many-PF} illustrates the Quality vs Efficiency of the minimalistic attacks. One objective of the Quality vs Efficiency is $100-Gen_{gt0}$ (to ensure the points closer to the bottom right corner are better), and the other is time cost in terms of wall clock time on completing 1000 function evaluations. 
For clarity, this figure displays only a subset of algorithms that are close to the frontier. It is clear that, across all cases, sTrEO and AMTEA significantly outperform in the first objective, while trailing behind some algorithms such as CGA and EKT in the second objective. The ``no free lunch theorem'' applies here, implying that no single algorithm will obtain better performance values than all the others in all problems.

\section{Conclusions}
\label{conclusion}

Recent studies have showcased the effectiveness of transfer evolutionary optimization (TrEO) in addressing complex problems. However, the ``no free lunch theorem'' (NFLT) in the TrEO literature underscores that no single algorithm can reign supreme across diverse problem types.
Remarkably, existing TrEO studies often focus on empirical analyses of optimization performance using synthetic benchmark functions, which often fall short due to inadequate design, predominantly featuring synthetic problems that lack real-world relevance. 
Therefore, this paper adopts a benchmarking approach to cultivate a deep understanding of the performance of TrEO algorithms, especially when confronted with diverse and complex transfer scenarios in practical problems. 

Specifically, we conducted a comprehensive analysis of three key characteristics with the advent of \textit{Big Source Task-Instances} in practical optimization problems, namely \textit{Big Volume}, \textit{Big Variety}, and \textit{Big Velocity}. We classified three representative problems including knapsack problems, planar robot arm problems, and minimalistic attacks. The knapsack problem represents a traditional discrete optimization problem characterized by \textit{Big Volume}. The planar arm problem is a continuous problem exhibiting both \textit{Big Volume} and \textit{Big Variety} characteristics. Lastly, minimalistic attacks encompass mixed problems that contain both \textit{Big Volume} and \textit{Big Velocity} characteristics.

Our work fills the gap by providing a practical benchmarking problem suite for researchers in the TrEO community. This initiative facilitates the assessment of algorithm performance on practical problems, promoting further interest and research in this promising field. It is important to note that our current benchmark suite includes only three types of practical test problems. Future research endeavors will focus on expanding the scope by incorporating additional benchmark problems with more diverse characteristics.

\ifCLASSOPTIONcaptionsoff
  \newpage
\fi

\bibliographystyle{IEEEtran}
\bibliography{IEEEabrv,references}

\begin{thebibliography}{10}
\providecommand{\url}[1]{#1}
\csname url@rmstyle\endcsname
\providecommand{\newblock}{\relax}
\providecommand{\bibinfo}[2]{#2}
\providecommand\BIBentrySTDinterwordspacing{\spaceskip=0pt\relax}
\providecommand\BIBentryALTinterwordstretchfactor{4}
\providecommand\BIBentryALTinterwordspacing{\spaceskip=\fontdimen2\font plus
\BIBentryALTinterwordstretchfactor\fontdimen3\font minus \fontdimen4\font\relax}
\providecommand\BIBforeignlanguage[2]{{%
\expandafter\ifx\csname l@#1\endcsname\relax
\typeout{** WARNING: IEEEtran.bst: No hyphenation pattern has been}%
\typeout{** loaded for the language `#1'. Using the pattern for}%
\typeout{** the default language instead.}%
\else
\language=\csname l@#1\endcsname
\fi
#2}}
\renewcommand\BIBentryALTinterwordstretchfactor{4}

\bibitem{ea-applications}
D.~Dasgupta and Z.~Michalewicz, \emph{Evolutionary algorithms in engineering applications}.\hskip 1em plus 0.5em minus 0.4em\relax Springer Science \& Business Media, 2013.

\bibitem{resource-constrained}
K.~Hindi, H.~Yang, and K.~Fleszar, ``An evolutionary algorithm for resource-constrained project scheduling,'' \emph{IEEE Transactions on Evolutionary Computation}, vol.~6, no.~5, pp. 512--518, 2002.

\bibitem{Resource-Investment-Project-Scheduling}
J.~Xiong, J.~Liu, Y.~Chen, and H.~A. Abbass, ``A knowledge-based evolutionary multiobjective approach for stochastic extended resource investment project scheduling problems,'' \emph{IEEE Transactions on Evolutionary Computation}, vol.~18, no.~5, pp. 742--763, 2014.

\bibitem{industrial-design}
A.~Benedetti, M.~Farina, and M.~Gobbi, ``Evolutionary multiobjective industrial design: the case of a racing car tire-suspension system,'' \emph{IEEE Transactions on Evolutionary Computation}, vol.~10, no.~3, pp. 230--244, 2006.

\bibitem{Satellite-Module-Layout-Design}
H.-f. Teng, Y.~Chen, W.~Zeng, Y.-j. Shi, and Q.-h. Hu, ``A dual-system variable-grain cooperative coevolutionary algorithm: Satellite-module layout design,'' \emph{IEEE Transactions on Evolutionary Computation}, vol.~14, no.~3, pp. 438--455, 2010.

\bibitem{cyber-physical-systems}
H.~He, C.~Maple, T.~Watson, A.~Tiwari, J.~Mehnen, Y.~Jin, and B.~Gabrys, ``The security challenges in the iot enabled cyber-physical systems and opportunities for evolutionary computing \& other computational intelligence,'' in \emph{2016 IEEE congress on evolutionary computation (CEC)}.\hskip 1em plus 0.5em minus 0.4em\relax IEEE, 2016, pp. 1015--1021.

\bibitem{cybersecurity}
J.~Kusyk, M.~U. Uyar, and C.~S. Sahin, ``Survey on evolutionary computation methods for cybersecurity of mobile ad hoc networks,'' \emph{Evolutionary Intelligence}, vol.~10, pp. 95--117, 2018.

\bibitem{cyber-attack-prevention}
D.~Zegzhda, D.~Lavrova, E.~Pavlenko, and A.~Shtyrkina, ``Cyber attack prevention based on evolutionary cybernetics approach,'' \emph{Symmetry}, vol.~12, no.~11, p. 1931, 2020.

\bibitem{Vehicle-Routing-Problem-With-Time-Windows}
P.~P. Repoussis, C.~D. Tarantilis, and G.~Ioannou, ``Arc-guided evolutionary algorithm for the vehicle routing problem with time windows,'' \emph{IEEE Transactions on Evolutionary Computation}, vol.~13, no.~3, pp. 624--647, 2009.

\bibitem{Two-Echelon-Vehicle-Routing}
X.~Yan, H.~Huang, Z.~Hao, and J.~Wang, ``A graph-based fuzzy evolutionary algorithm for solving two-echelon vehicle routing problems,'' \emph{IEEE Transactions on Evolutionary Computation}, vol.~24, no.~1, pp. 129--141, 2020.

\bibitem{Evolutionary-Machine-Learning}
M.~A. Franco, N.~Krasnogor, and J.~Bacardit, ``Automatic tuning of rule-based evolutionary machine learning via problem structure identification,'' \emph{IEEE Computational Intelligence Magazine}, vol.~15, no.~3, pp. 28--46, 2020.

\bibitem{Evolutionary-Automated-Machine-Learning}
M.~Sarafanov, V.~Pokrovskii, and N.~O. Nikitin, ``Evolutionary automated machine learning for multi-scale decomposition and forecasting of sensor time series,'' in \emph{2022 IEEE Congress on Evolutionary Computation (CEC)}.\hskip 1em plus 0.5em minus 0.4em\relax IEEE, 2022, pp. 01--08.

\bibitem{CIGAR}
S.~J. Louis and J.~McDonnell, ``Learning with case-injected genetic algorithms,'' \emph{IEEE Transactions on Evolutionary Computation}, vol.~8, no.~4, pp. 316--328, 2004.

\bibitem{air5}
Y.-S. Ong and A.~Gupta, ``Air 5: Five pillars of artificial intelligence research,'' \emph{IEEE Transactions on Emerging Topics in Computational Intelligence}, vol.~3, no.~5, pp. 411--415, 2019.

\bibitem{insights}
A.~Gupta, Y.-S. Ong, and L.~Feng, ``Insights on transfer optimization: Because experience is the best teacher,'' \emph{IEEE Transactions on Emerging Topics in Computational Intelligence}, vol.~2, no.~1, pp. 51--64, 2017.

\bibitem{revies-on-mto}
T.~Wei, S.~Wang, J.~Zhong, D.~Liu, and J.~Zhang, ``A review on evolutionary multitask optimization: Trends and challenges,'' \emph{IEEE Transactions on Evolutionary Computation}, vol.~26, no.~5, pp. 941--960, 2021.

\bibitem{half-a-dozen}
A.~Gupta, L.~Zhou, Y.-S. Ong, Z.~Chen, and Y.~Hou, ``Half a dozen real-world applications of evolutionary multitasking, and more,'' \emph{IEEE Computational Intelligence Magazine}, vol.~17, no.~2, pp. 49--66, 2022.

\bibitem{EKT}
E.~O. Scott and K.~A. De~Jong, ``First complexity results for evolutionary knowledge transfer,'' in \emph{Proceedings of the 17th ACM/SIGEVO Conference on Foundations of Genetic Algorithms}, 2023, pp. 140--151.

\bibitem{multitasking-single-benchmark}
B.~Da, Y.-S. Ong, L.~Feng, A.~K. Qin, A.~Gupta, Z.~Zhu, C.-K. Ting, K.~Tang, and X.~Yao, ``Evolutionary multitasking for single-objective continuous optimization: Benchmark problems, performance metric, and baseline results,'' \emph{arXiv preprint arXiv:1706.03470}, 2017.

\bibitem{multitasking-multi-benchmark}
Y.~Yuan, Y.-S. Ong, L.~Feng, A.~K. Qin, A.~Gupta, B.~Da, Q.~Zhang, K.~C. Tan, Y.~Jin, and H.~Ishibuchi, ``Evolutionary multitasking for multiobjective continuous optimization: Benchmark problems, performance metrics and baseline results,'' \emph{arXiv preprint arXiv:1706.02766}, 2017.

\bibitem{cec2020benchmark}
A.~Viktorin, R.~Senkerik, M.~Pluhacek, T.~Kadavy, and A.~Zamuda, ``Dish-xx solving cec2020 single objective bound constrained numerical optimization benchmark,'' in \emph{2020 IEEE Congress on Evolutionary Computation (CEC)}, 2020, pp. 1--8.

\bibitem{ExTrEMO}
J.~Liu, A.~Gupta, C.~Ooi, and Y.-S. Ong, ``Extremo: Transfer evolutionary multiobjective optimization with proof of faster convergence,'' \emph{IEEE Transactions on Evolutionary Computation}, pp. 1--1, 2024.

\bibitem{coping-with-big-data-MAB-AMTEA}
M.~Shakeri, A.~Gupta, Y.-S. Ong, X.~Chi, and A.~Z. NengSheng, ``Coping with big data in transfer optimization,'' in \emph{2019 IEEE International Conference on Big Data (Big Data)}.\hskip 1em plus 0.5em minus 0.4em\relax IEEE, 2019, pp. 3925--3932.

\bibitem{sTrEO}
M.~Shakeri, E.~Miahi, A.~Gupta, and Y.~Ong, ``Scalable transfer evolutionary optimization: Coping with big task instances,'' \emph{IEEE transactions on cybernetics}, vol.~PP, 2022.

\bibitem{domain-adaptation}
R.~Lim, A.~Gupta, Y.-S. Ong, L.~Feng, and A.~N. Zhang, ``Non-linear domain adaptation in transfer evolutionary optimization,'' \emph{Cognitive Computation}, vol.~13, pp. 290--307, 2021.

\bibitem{memetic-computation:}
A.~Gupta and Y.-S. Ong, \emph{Memetic computation: the mainspring of knowledge transfer in a data-driven optimization era}.\hskip 1em plus 0.5em minus 0.4em\relax Springer, 2018, vol.~21.

\bibitem{information-geometric}
Y.~Ollivier, L.~Arnold, A.~Auger, and N.~Hansen, ``Information-geometric optimization algorithms: A unifying picture via invariance principles,'' \emph{The Journal of Machine Learning Research}, vol.~18, no.~1, pp. 564--628, 2017.

\bibitem{ETO}
K.~C. Tan, L.~Feng, and M.~Jiang, ``Evolutionary transfer optimization-a new frontier in evolutionary computation research,'' \emph{IEEE Computational Intelligence Magazine}, vol.~16, no.~1, pp. 22--33, 2021.

\bibitem{neuro-evolution}
J.~C. Wong, A.~Gupta, and Y.-S. Ong, ``Can transfer neuroevolution tractably solve your differential equations?'' \emph{IEEE Computational Intelligence Magazine}, vol.~16, no.~2, pp. 14--30, 2021.

\bibitem{multiproblem}
A.~T.~W. Min, Y.-S. Ong, A.~Gupta, and C.-K. Goh, ``Multiproblem surrogates: Transfer evolutionary multiobjective optimization of computationally expensive problems,'' \emph{IEEE Transactions on Evolutionary Computation}, vol.~23, no.~1, pp. 15--28, 2017.

\bibitem{objective-heterogeneous}
X.~Xue, C.~Yang, Y.~Hu, K.~Zhang, Y.-M. Cheung, L.~Song, and K.~C. Tan, ``Evolutionary sequential transfer optimization for objective-heterogeneous problems,'' \emph{IEEE Transactions on Evolutionary Computation}, vol.~26, no.~6, pp. 1424--1438, 2021.

\bibitem{population-based}
S.~Yang and X.~Yao, ``Population-based incremental learning with associative memory for dynamic environments,'' \emph{IEEE Transactions on Evolutionary Computation}, vol.~12, no.~5, pp. 542--561, 2008.

\bibitem{direct-memory-schemes}
M.~Mavrovouniotis and S.~Yang, ``Direct memory schemes for population-based incremental learning in cyclically changing environments,'' in \emph{Applications of Evolutionary Computation: 19th European Conference, EvoApplications 2016, Porto, Portugal, March 30--April 1, 2016, Proceedings, Part II 19}, vol. 9598.\hskip 1em plus 0.5em minus 0.4em\relax Springer, 2016, pp. 233--247.

\bibitem{memetic-search}
L.~Feng, Y.-S. Ong, M.-H. Lim, and I.~W. Tsang, ``Memetic search with interdomain learning: A realization between cvrp and carp,'' \emph{IEEE Transactions on Evolutionary Computation}, vol.~19, no.~5, pp. 644--658, 2014.

\bibitem{memes-as-building-blocks}
L.~Feng, Y.-S. Ong, A.-H. Tan, and I.~W. Tsang, ``Memes as building blocks: a case study on evolutionary optimization+ transfer learning for routing problems,'' \emph{Memetic Computing}, vol.~7, pp. 159--180, 2015.

\bibitem{transfer-learning-in-gp}
T.~T.~H. Dinh, T.~H. Chu, and Q.~U. Nguyen, ``Transfer learning in genetic programming,'' in \emph{2015 IEEE Congress on Evolutionary Computation (CEC)}.\hskip 1em plus 0.5em minus 0.4em\relax IEEE, 2015, pp. 1145--1151.

\bibitem{further-investigation}
E.~Haslam, B.~Xue, and M.~Zhang, ``Further investigation on genetic programming with transfer learning for symbolic regression,'' in \emph{2016 IEEE Congress on Evolutionary Computation (CEC)}.\hskip 1em plus 0.5em minus 0.4em\relax IEEE, 2016, pp. 3598--3605.

\bibitem{extracting}
M.~Iqbal, W.~N. Browne, and M.~Zhang, ``Extracting and using building blocks of knowledge in learning classifier systems,'' in \emph{Proceedings of the 14th annual conference on Genetic and evolutionary computation}.\hskip 1em plus 0.5em minus 0.4em\relax Association for Computing Machinery, 2012, pp. 863--870.

\bibitem{reusing-building-blocks}
M.~Iqbal, W.~N. Browne, and M.~Zhang, ``Reusing building blocks of extracted knowledge to solve complex, large-scale boolean problems,'' \emph{IEEE Transactions on Evolutionary Computation}, vol.~18, no.~4, pp. 465--480, 2013.

\bibitem{evolutionary-robotics}
A.~Moshaiov and A.~Tal, ``Family bootstrapping: A genetic transfer learning approach for onsetting the evolution for a set of related robotic tasks,'' in \emph{2014 IEEE Congress on Evolutionary Computation (CEC)}.\hskip 1em plus 0.5em minus 0.4em\relax IEEE, 2014, pp. 2801--2808.

\bibitem{cross-domain-reuse}
M.~Iqbal, B.~Xue, H.~Al-Sahaf, and M.~Zhang, ``Cross-domain reuse of extracted knowledge in genetic programming for image classification,'' \emph{IEEE Transactions on Evolutionary Computation}, vol.~21, no.~4, pp. 569--587, 2017.

\bibitem{frame-correlation}
X.~Qu, Y.-S. Ong, and A.~Gupta, ``Frame-correlation transfers trigger economical attacks on deep reinforcement learning policies,'' \emph{IEEE Transactions on Cybernetics}, vol.~52, no.~8, pp. 7577--7590, 2022.

\bibitem{genetic-transfer-learning}
B.~Ko{\c{c}}er and A.~Arslan, ``Genetic transfer learning,'' \emph{Expert Systems with Applications}, vol.~37, no.~10, pp. 6997--7002, 2010.

\bibitem{curbing-AMTEA}
B.~Da, A.~Gupta, and Y.-S. Ong, ``Curbing negative influences online for seamless transfer evolutionary optimization,'' \emph{IEEE Transactions on Cybernetics}, vol.~49, no.~12, pp. 4365--4378, 2018.

\bibitem{MFEA-DV}
J.~Yin, A.~Zhu, Z.~Zhu, Y.~Yu, and X.~Ma, ``Multifactorial evolutionary algorithm enhanced with cross-task search direction,'' in \emph{2019 IEEE Congress on Evolutionary Computation (CEC)}, 2019, pp. 2244--2251.

\bibitem{MSSTO}
J.~Zhang, W.~Zhou, X.~Chen, W.~Yao, and L.~Cao, ``Multisource selective transfer framework in multiobjective optimization problems,'' \emph{IEEE Transactions on Evolutionary Computation}, vol.~24, no.~3, pp. 424--438, 2020.

\bibitem{KAES}
L.~Zhou, L.~Feng, A.~Gupta, and Y.-S. Ong, ``Learnable evolutionary search across heterogeneous problems via kernelized autoencoding,'' \emph{IEEE Transactions on Evolutionary Computation}, vol.~25, no.~3, pp. 567--581, 2021.

\bibitem{AT-MFEA}
X.~Xue, K.~Zhang, K.~C. Tan, L.~Feng, J.~Wang, G.~Chen, X.~Zhao, L.~Zhang, and J.~Yao, ``Affine transformation-enhanced multifactorial optimization for heterogeneous problems,'' \emph{IEEE Transactions on Cybernetics}, vol.~52, no.~7, pp. 6217--6231, 2022.

\bibitem{MSTL-DMOEA}
Y.~Ye, Q.~Lin, L.~Ma, K.-C. Wong, M.~Gong, and C.~A.~C. Coello, ``Multiple source transfer learning for dynamic multiobjective optimization,'' \emph{Information Sciences}, vol. 607, pp. 739--757, 2022.

\bibitem{CIGAR-application}
S.~J. Louis and C.~Miles, ``Playing to learn: Case-injected genetic algorithms for learning to play computer games,'' \emph{IEEE Transactions on Evolutionary Computation}, vol.~9, no.~6, pp. 669--681, 2005.

\bibitem{minimalistic-attack}
Q.~Xinghua, Z.~Sun, Y.~Ong, A.~Gupta, and P.~Wei, ``Minimalistic attacks: How little it takes to fool deep reinforcement learning policies,'' \emph{IEEE Transactions on Cognitive and Developmental Systems}, vol.~PP, pp. 1--1, 02 2020.

\bibitem{KBOA}
J.~Schwarz and J.~Ocenasek, ``A problem knowledge-based evolutionary algorithm kboa for hypergraph bisectioning,'' in \emph{Proceedings of the 4th joint conference on knowledge-based software engineering. IOS Press}, 2000, pp. 51--58.

\bibitem{MOI-MBO}
B.~Bischl, S.~Wessing, N.~Bauer, K.~Friedrichs, and C.~Weihs, ``Moi-mbo: multiobjective infill for parallel model-based optimization,'' in \emph{Learning and Intelligent Optimization: 8th International Conference, Lion 8, Gainesville, FL, USA, February 16-21, 2014. Revised Selected Papers 8}, vol. 8426.\hskip 1em plus 0.5em minus 0.4em\relax Springer, 2014, pp. 173--186.

\bibitem{solution-representation}
R.~Lim, L.~Zhou, A.~Gupta, Y.-S. Ong, and A.~N. Zhang, ``Solution representation learning in multi-objective transfer evolutionary optimization,'' \emph{IEEE Access}, vol.~9, pp. 41\,844--41\,860, 2021.

\bibitem{to-handle-big-data}
G.~Sanchita and D.~Anindita, ``Evolutionary algorithm based techniques to handle big data,'' \emph{Techniques and Environments for Big Data Analysis: Parallel, Cloud, and Grid Computing}, pp. 113--158, 2016.

\bibitem{real-time-optimization}
M.~Diehl, H.~G. Bock, J.~P. Schl{\"o}der, R.~Findeisen, Z.~Nagy, and F.~Allg{\"o}wer, ``Real-time optimization and nonlinear model predictive control of processes governed by differential-algebraic equations,'' \emph{Journal of Process Control}, vol.~12, no.~4, pp. 577--585, 2002.

\bibitem{knapsack-problems:}
S.~Martello and P.~Toth, \emph{Knapsack problems: algorithms and computer implementations}.\hskip 1em plus 0.5em minus 0.4em\relax John Wiley \& Sons, Inc., 1990.

\bibitem{10.1007/3-540-58495-1_14}
Z.~Michalewicz and J.~Arabas, ``Genetic algorithms for the 0/1 knapsack problem,'' in \emph{Methodologies for Intelligent Systems}, Z.~W. Ra{\'{s}} and M.~Zemankova, Eds., vol. 869.\hskip 1em plus 0.5em minus 0.4em\relax Berlin, Heidelberg: Springer Berlin Heidelberg, 1994, pp. 134--143.

\bibitem{quality-diversity}
J.-B. Mouret and G.~Maguire, ``Quality diversity for multi-task optimization,'' in \emph{Proceedings of the 2020 Genetic and Evolutionary Computation Conference}.\hskip 1em plus 0.5em minus 0.4em\relax Association for Computing Machinery, 2020, pp. 121--129.

\bibitem{cga}
J.~Holland, \emph{{Adaptation in natural and artificial systems}}.\hskip 1em plus 0.5em minus 0.4em\relax The University of Michigan Press, Ann Arbor, 1975.

\bibitem{mvarnorm}
C.~B. Do, ``The multivariate gaussian distribution,'' \emph{Section Notes, Lecture on Machine Learning, CS}, vol. 229, 2008.

\bibitem{uniform-crossover}
G.~Syswerda \emph{et~al.}, ``Uniform crossover in genetic algorithms.'' in \emph{ICGA}, vol.~3, 1989, pp. 2--9.

\bibitem{umd}
H.~Mühlenbein, ``The equation for response to selection and its use for prediction,'' \emph{Evolutionary Computation}, vol.~5, no.~3, pp. 303--346, 1997.

\bibitem{SBX}
K.~Deb, R.~B. Agrawal, \emph{et~al.}, ``Simulated binary crossover for continuous search space,'' \emph{Complex systems}, vol.~9, no.~2, pp. 115--148, 1995.

\bibitem{crossover-operators}
P.~Kora and P.~Yadlapalli, ``Crossover operators in genetic algorithms: A review,'' \emph{International Journal of Computer Applications}, vol. 162, no.~10, 2017.

\bibitem{polynomial-mutation}
K.~Deb, D.~Deb, \emph{et~al.}, ``Analysing mutation schemes for real-parameter genetic algorithms.'' \emph{Int. J. Artif. Intell. Soft Comput.}, vol.~4, no.~1, pp. 1--28, 2014.

\bibitem{autoencoding}
L.~Feng, Y.-S. Ong, S.~Jiang, and A.~Gupta, ``Autoencoding evolutionary search with learning across heterogeneous problems,'' \emph{IEEE Transactions on Evolutionary Computation}, vol.~21, no.~5, pp. 760--772, 2017.

\bibitem{acktr}
Y.~Wu, E.~Mansimov, R.~B. Grosse, S.~Liao, and J.~Ba, ``Scalable trust-region method for deep reinforcement learning using kronecker-factored approximation,'' \emph{Advances in neural information processing systems}, vol.~30, 2017.

\end{thebibliography}

\vfill

\end{document}